%% file: main.tex
\documentclass[10pt,journal,compsoc]{IEEEtran}

\usepackage{enumitem}

\ifCLASSOPTIONcompsoc
  \usepackage[nocompress]{cite}
\else
  \usepackage{cite}
\fi
\usepackage{graphicx} 
\ifCLASSINFOpdf
\else
\fi
\usepackage{amsmath}
\usepackage{mathrsfs}
\usepackage{euscript}
\usepackage{acronym}
\usepackage{algorithmic}

\usepackage{mdwmath}
\usepackage{mdwtab}

\usepackage{amssymb}
\usepackage{booktabs}
\usepackage{multirow}
\usepackage{graphicx}
\usepackage{subfigure}
\usepackage{color}
\usepackage{xcolor}
\ifCLASSOPTIONcompsoc
  \usepackage[caption=false,font=footnotesize,labelfont=sf,textfont=sf]{subfig}
\else
  \usepackage[caption=false,font=footnotesize]{subfig}
\fi
\usepackage{caption}
\usepackage{url} 
\begin{document}
%
\title{Paint Bucket Colorization Using Anime Character Color Design Sheets}
%
%
%
%

\author{Yuekun~Dai, Qinyue~Li,
        Shangchen~Zhou,
        Yihang~Luo,
        Chongyi~Li, 
        Chen~Change~Loy,~\IEEEmembership{Senior~Member,~IEEE}
\IEEEcompsocitemizethanks{
\IEEEcompsocthanksitem Y. Dai, Q. Li, S. Zhou, Y. Luo, and C. C. Loy are with S-Lab, Nanyang Technological  University (NTU), Singapore. C. Li is with Nankai University, China. (E-mail: ydai005@e.ntu.edu.sg, qinyue.li@ntu.edu.sg, sczhou@ntu.edu.sg, c200211@e.ntu.edu.sg, ccloy@ntu.edu.sg, lichongyi@nankai.edu.cn) 
\protect\\
\IEEEcompsocthanksitem C. C. Loy is the corresponding author.}
}

%
%

\markboth{}
{Shell \MakeLowercase{\textit{et al.}}: Paint Bucket Colorization Using Anime Character Color Design Sheets}

%


\newcommand{\pami}[1]{{\color{black}{#1}}}

\newcommand{\lichongyi}[1]{\textbf{\color{red}(CY: {#1})}}
\newcommand{\shangchen}[1]{\textbf{\color{red}(SC: {#1})}}
\newcommand{\yuekun}[1]{\textbf{\color{green}(YK: {#1})}}
\newcommand{\cavan}[1]{\textbf{\color{cyan}(CV: {#1})}}
\newcommand{\rebuttal}[1]{{\color{black}{#1}}}
\newcommand{\dataset}{\textit{\text{PaintBucket}}}
\newcommand{\pbc}{\textit{PaintBucket-Character}}

\newcommand{\eg}{\emph{e.g.,}}
\newcommand{\etal}{\emph{et al.}}

\newcommand{\colorbook}{colorbook}




\input{sec/0_abstract}
\maketitle
\IEEEdisplaynontitleabstractindextext

%
\IEEEpeerreviewmaketitle


\begin{figure*}
  \vspace{-3mm}
  \includegraphics[width=\textwidth]{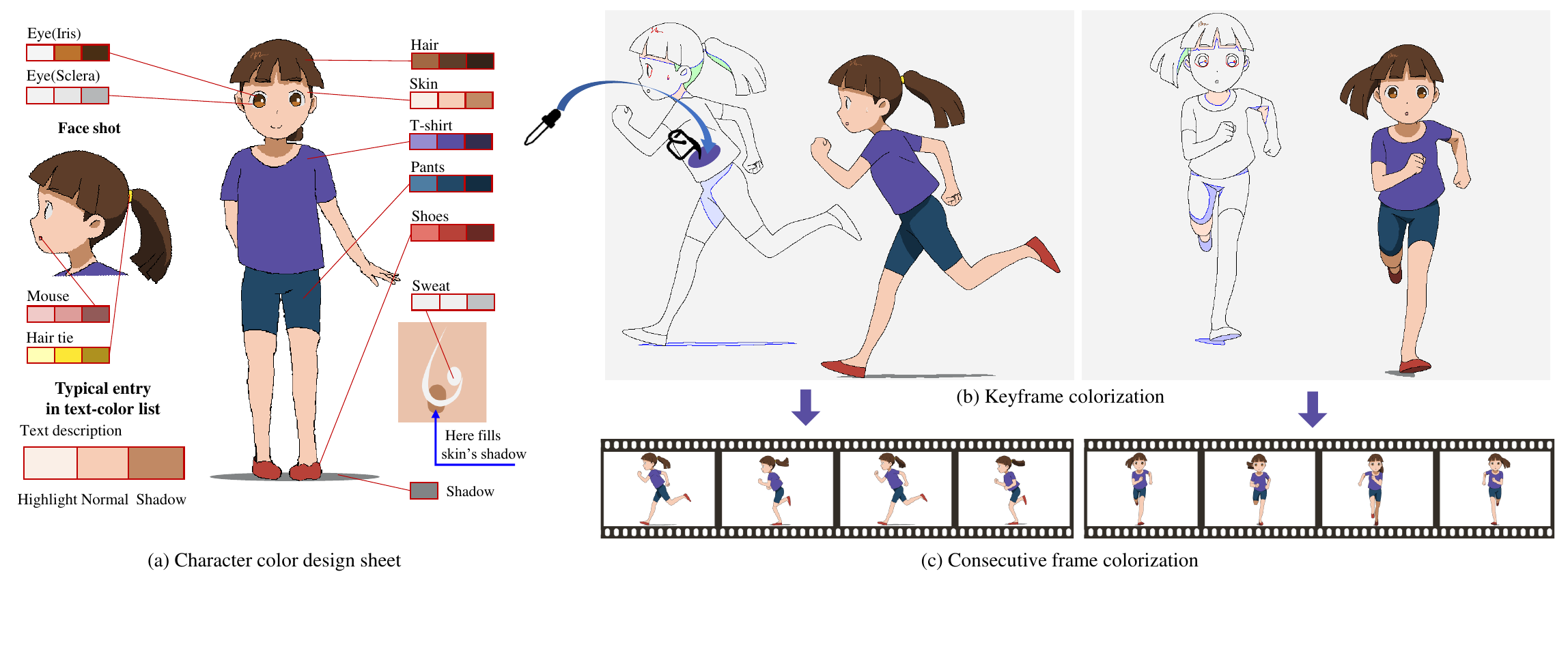}
  \vspace{-14mm}
  \caption{\pami{
  In animation production, digital painters refer to the character color design sheet and use the paint bucket tool to colorize drawn line art frame by frame. This character color design sheet includes a text-color list, which provides a text description along with highlight, normal, and shadow colors for each semantic element of the character.
  To automate this process, we propose a \textit{keyframe colorization} method and a \textit{consecutive frames colorization} method.
  Our keyframe colorization approach can colorize the keyframe line arts in different poses and perspectives based on a given character color design sheet.
  Then, the consecutive frames colorization method can propagate these colors throughout the sequence to achieve automatic paint bucket colorization.
  \copyright{Drawn by NeonRed, used with artist's permission.}}
  }
  \label{fig:teaser}
  \vspace{-5mm}
\end{figure*}

\input{sec/1_intro}

\input{sec/2_related_work}
\input{sec/3_dataset}
\input{sec/4_method}
\input{sec/5_experiments}
\input{sec/6_limitation}
\input{sec/7_conclusion}

\ifCLASSOPTIONcaptionsoff
  \newpage
\fi



%

\medskip
{
\small
\bibliographystyle{IEEEtran}
\bibliography{reference}
}

\begin{IEEEbiography}
[{\includegraphics[width=1in,height=1.25in,clip,keepaspectratio]{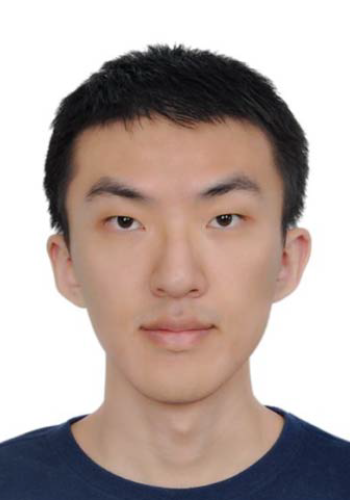}}]
{Yuekun Dai}
is currently a second-year PhD student at MMLab@NTU, Nanyang Technological University. He received his M.S. degree in Nanyang Technological University as well as his B.S. degree from Peking University. 
He co-organized the ``MIPI workshop'' series in conjunction with CVPR 2023 and CVPR 2024, and was selected as an outstanding reviewer in BMVC 2022.
His research paper on nighttime flare removal was accepted by CVPR 2023 and selected as a highlight.
His research interests include computational photography and computer graphics.
\end{IEEEbiography}

\begin{IEEEbiography}[{\includegraphics[width=1in,height=1.25in,clip,keepaspectratio]{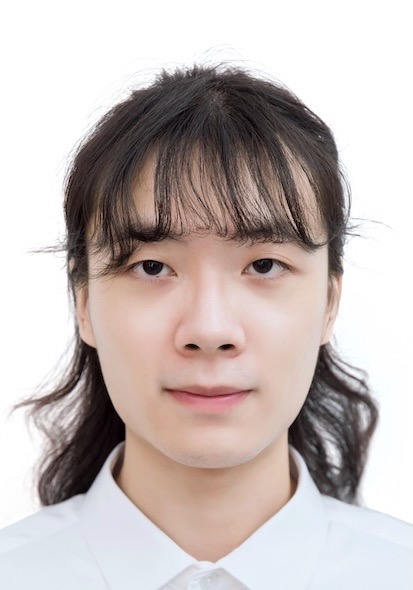}}]
{Qinyue Li} is currently a research assistant at S-Lab, Nanyang Technological University, Singapore. He graduated with a master's degree from the National University of Singapore and a bachelor's degree from the University of Electronic Science and Technology of China. His research interests are in computer vision, especially computational photography.
\vspace{-4mm}
\end{IEEEbiography}

\begin{IEEEbiography}[{\includegraphics[width=1in,height=1.25in,clip,keepaspectratio]{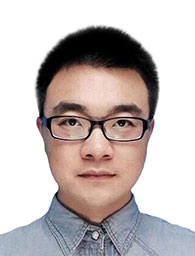}}]
{Shangchen Zhou} is currently a Research Fellow at MMLab@NTU, Nanyang Technological University, Singapore. He received his Ph.D. (2024) in Computer Science from the same institution. He was selected as an outstanding reviewer in NeurIPS 2020. He won first place in three image restoration and enhancement challenges in NTIRE 2021. His works received notable recognition including the WAIC Youth Outstanding Paper Award Honorable Mention in 2023, the Snap Fellowship Honorable Mention in 2022, and the Best Paper Award at ICIMCS, ACM in 2016. Additionally, He co-organized the ``MIPI workshop'' series in conjunction with ECCV 2022, CVPR 2023, and CVPR 2024. His research interests include image/video restoration and enhancement, generation and editing, etc.
\vspace{-4mm}
\end{IEEEbiography}

\begin{IEEEbiography}
[{\includegraphics[width=1in,height=1.25in,clip,keepaspectratio]{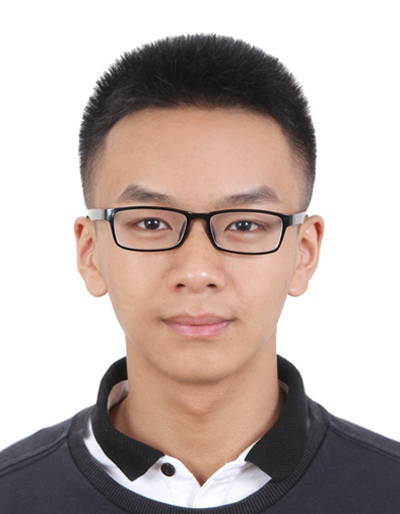}}]
{Yihang Luo}
is currently a Research Engineer MMLab@NTU, Nanyang Technological University, Singapore, specializing in Computer Science. He serves as an undergraduate research assistant at MMLab@NTU, S-Lab. He was awarded the Singapore Science and Engineering Scholarship from 2020 to 2024, and the NTU President Research Scholarship from 2021 to 2023. His research interests encompass computer vision and machine learning, particularly in image/video restoration and enhancement.
\vspace{-4mm}
\end{IEEEbiography}

\begin{IEEEbiography}
[{\includegraphics[width=1in,height=1.25in,clip,keepaspectratio]{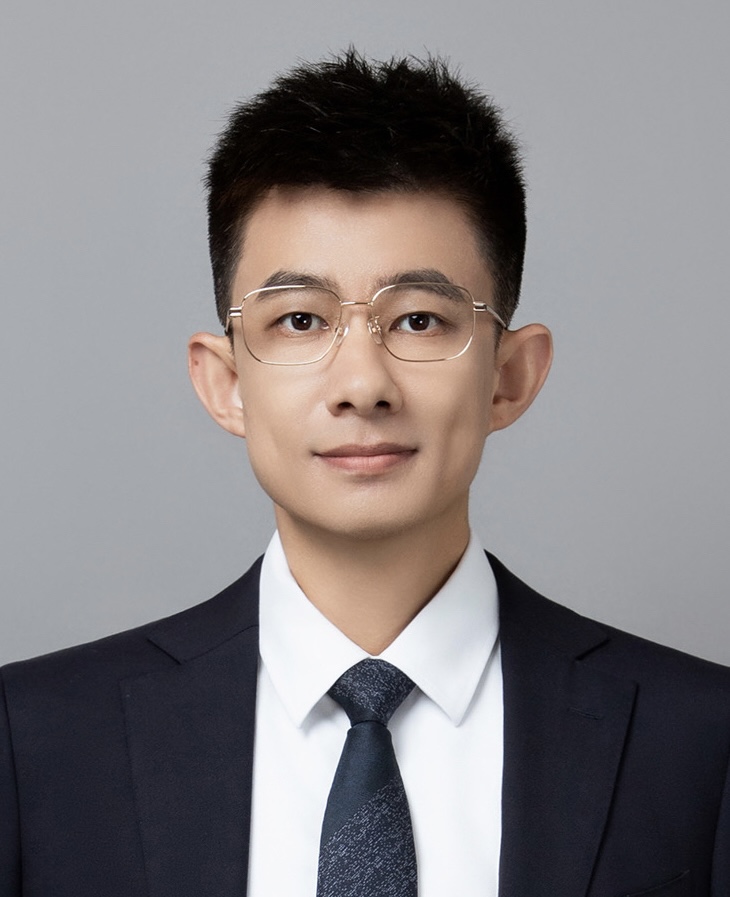}}]
{Chongyi Li} is a Professor at the School of Computer Science, Nankai University, China.  He was a  Research Assistant Professor with the MMLab@NTU, S-Lab,  School of Computer Science and Engineering, Nanyang Technological University, Singapore from 2021 to 2023.  He was a Research Fellow with the City University of Hong Kong and Nanyang Technological University from 2018 to 2021. He received his Ph.D. degree from Tianjin University in 2018. He was also a joint training Ph.D. student at the Australian National University, Australia, from 2016 to 2017. His research interests include computer vision, machine learning, and computational imaging, particularly in image enhancement and restoration, image generation and editing, and underwater imaging. 
\vspace{-4mm}
\end{IEEEbiography}

\begin{IEEEbiography}
[{\includegraphics[width=1in,height=1.25in,clip,keepaspectratio]{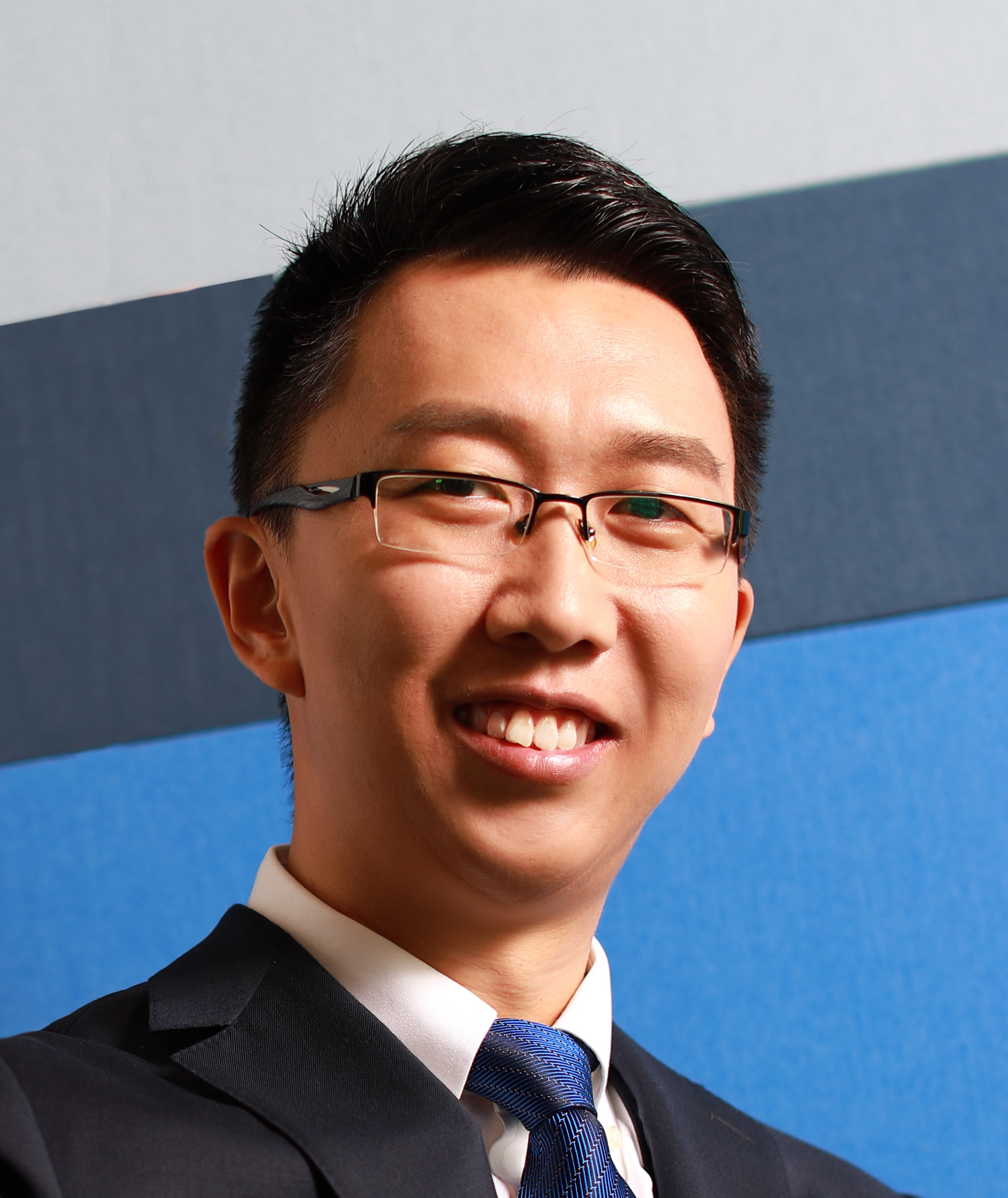}}]{Chen Change Loy}
(Senior Member, IEEE) is a President's Chair Professor at the School of Computer Science and Engineering, Nanyang Technological University, Singapore. Before joining NTU, he served as a Research Assistant Professor at the MMLab of The Chinese University of Hong Kong, from 2013 to 2018. He received his Ph.D. (2010) in Computer Science from the Queen Mary University of London. He was a postdoctoral researcher at Queen Mary University of London and Vision Semantics Limited, from 2010 to 2013.
He serves/served as an Associate Editor of the IEEE Transactions on Pattern Analysis and Machine Intelligence, International Journal of Computer Vision and Computer Vision and Image Understanding. He also serves/served as an Area Chair of major conferences such as ICCV, CVPR, ECCV, NeurIPS, and ICLR. He is a senior member of IEEE.
His research interests include image/video restoration and enhancement, generative tasks, and representation learning.
\end{IEEEbiography}




\end{document}

%% file: sec/0_abstract.tex
\IEEEtitleabstractindextext{%
\begin{abstract}
\pami{
Line art colorization plays a crucial role in hand-drawn animation production, where digital artists manually colorize segments using a paint bucket tool, guided by RGB values from character color design sheets. This process, often called paint bucket colorization, involves two main tasks: keyframe colorization, where colors are applied according to the character's color design sheet, and consecutive frame colorization, where these colors are replicated across adjacent frames.
Current automated colorization methods primarily focus on reference-based and segment-matching approaches. However, reference-based methods often fail to accurately assign specific colors to each region, while matching-based methods are limited to consecutive frame colorization and struggle with issues like significant deformation and occlusion.
In this work, we introduce inclusion matching, which allows the network to understand the inclusion relationships between segments, rather than relying solely on direct visual correspondences. By integrating this approach with segment parsing and color warping modules, our inclusion matching pipeline significantly improves performance in both keyframe colorization and consecutive frame colorization.
To support our network's training, we have developed a unique dataset named \pbc, which includes rendered line arts alongside their colorized versions and shading annotations for various 3D characters. To replicate industry animation data formats, we also created color design sheets for each character, with semantic description for each color and standard pose reference images.
Experiments highlight the superiority of our method, demonstrating accurate and consistent colorization across both our proposed benchmarks and hand-drawn animations.
Data and code are available at \url{https://github.com/ykdai/BasicPBC}.
}
\vspace{-4mm}
\end{abstract}

}

%% file: sec/1_intro.tex
\ifCLASSOPTIONcompsoc
\IEEEraisesectionheading{\section{Introduction}\label{sec:intro}}
\else
\section{Introduction}
\label{sec:intro}
\fi

\pami{

\IEEEPARstart{C}olorizing line art is an important step in animation production. In commercial animation, anime character color design sheets, such as the one shown in Fig.~\ref{fig:teaser}(a), are created to guide digital painters in the colorization process. These sheets provide a text-color list for all regions with different semantic elements of each character.
A typical entry in the text-color list consists of a text description and three specific colors corresponding to each semantic element: a highlight color, a normal color, and a shadow color. To ensure clarity for digital painters, animators use color lines and pre-colorized segments to mark highlight and shadow regions, as illustrated in Fig.~\ref{fig:teaser}(b).
Following these annotations and the character design sheet, digital painters use the paint bucket tool to colorize each line-enclosed segment and the corresponding color line in the keyframe. They then proceed to colorize every frame in the sequence based on the keyframe's colors. These two processes, known as \textit{keyframe colorization} and \textit{consecutive frame colorization}, are collectively referred to as paint bucket colorization~\cite{InclusionMatching2024}.

Reference-based colorization and segment matching are two primary approaches for paint bucket colorization.
However, reference-based colorization methods~\cite{sykora2004unsupervised,sykora2009rigid,chen2020active, lee2020reference,wu2023self,zhang2021line,controlnet,akita2023hand,cao2023animediffusion,xing2024tooncrafter} cannot produce accurate colors predefined in the color design sheet for each line-enclosed segment.
Segment matching methods~\cite{AnT,siyao2022animerun,maejima2019graph,maejima2024continual} aim to match similar segment across frames based on segments' shape similarities and relative positions, but they encounter significant challenges in scenarios involving occlusion or large motion.
In these cases, motion disrupts segment similarity, and fragmentation breaks the one-to-one correspondence between segments, making segment matching inherently problematic and poorly defined.

In this paper, we introduce the novel notion of ``{inclusion matching}'', which bypasses the necessity for exact segment-to-segment correspondence.
Inclusion matching aims to compute the likelihood of each segment in a target frame being included within a specific color in the reference frame.
Building on inclusion matching, we revisit the industrial anime production pipeline and propose the first color-design-sheet-based colorization pipeline, featuring the \textit{keyframe colorization} and \textit{consecutive frame colorization}.
These methods can be seamlessly integrated into the current animation production pipeline.

\textit{Keyframe colorization} based on a color design sheet requires both semantic segmentation and detailed matching. 
For regions with clear semantic information, like hair, semantic guidance is essential, while precise matching is paramount for areas with numerous small details, such as a row of buttons in different colors.
Achieving accurate keyframe colorization is challenging.
Initially, the reference frame and the target frame always consist of different types of data. The reference frame includes fully colorized line art and text descriptions for each color, whereas the target frame only provides line art with annotated regions for shadows and highlights. Connecting the corresponding segments with different data modalities is difficult.
Furthermore, many regions with similar shapes and close proximity, such as the collar and the neck, are inherently confusing.
To address these issues, we propose a paint bucket colorization pipeline with a segment parsing module and an inclusion matching module.
We categorize text descriptions of common parts of the character into 12 semantic categories, containing background, bag, belt, glasses, hair, socks, hat, mouth, clothes, eyes, shoes, and skin.
Given the different modalities of reference and target frames, the segment parsing module uses contrastive learning to align the target frame's image features with the semantic categories in the reference.
For confusing regions, animators are required to clarify these areas with specific annotations in the animation production specification~\cite{kyoto_draw}.
These animators use different colors to mark highlights and shadows on the skin, hair, and other regions, which we refer to as shading annotations, as illustrated in Fig.~\ref{fig:ani_production}(c).
Inspired by this, we incorporate these annotations as inputs to our system to avoid confusing situations.
Then, the inclusion matching module match the details (\eg~the iris of the eyes) across reference and target frames within the same parsing.
By leveraging these strategies, our approach effectively improves the accuracy of reference-based keyframe colorization.

\begin{figure*}[t]
  \vspace{-3mm}
  \includegraphics[width=\textwidth]{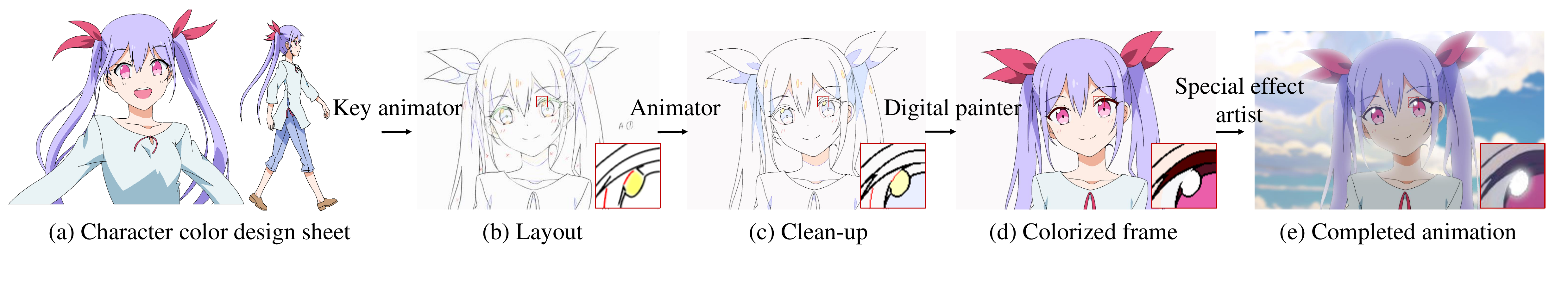}
  \vspace{-13mm}
  \caption{\pami{Production procedure of the hand-drawn animation. Animators pre-colorize regions such as highlights and shadows while using different colors to differentiate regions of hair, skin, and others in the clean-up stage. These colors, known as shading annotations, serve as guides for digital painters, helping prevent confusion during the coloring process. \copyright{drawn by Nicca (Sriprachum Kongwisawamit), used with artist permission.}}}
  \vspace{-6mm}
  \label{fig:ani_production}
\end{figure*}


For \textit{consecutive frame colorization}, we follow a colorization pipeline that includes a color warping module and an inclusion matching module, inspired by the approach of skilled digital painters. 
This approach typically begins with coloring larger, more noticeable segments across all frames before filling in the details of smaller segments in each frame.
Motivated by this workflow, our pipeline first computes optical flow between line sketches, and warp colors from a reference (keyframe) to a target frame to achieve preliminary and coarse colorization. 
This is followed by the application of an inclusion matching module, which refines the colorization result of tiny segments.
Our \textit{consecutive frame colorization} allows us to propagate the colors from the keyframe to subsequent frames, completing the colorization of the entire line art sequence.

Dataset plays a key role in the success of our method. To train and evaluate our method, we select 22 character models from Mixamo~\cite {mixamo} and Aplaybox~\cite{aplaybox}. 
Then, we use 3D software Cinema 4D to synthesize a specially-designed cartoon character colorization dataset named \pbc, mimicking real intermediate line art data on the animation colorization production and its colorized ground truth. 
We disable the anti-aliasing for our rendered data and provide the color lines to satisfy animation production standard.
Since there is no existing paint bucket dataset or benchmark that contains the character's color design sheet and shading annotations, we label the text description of character's each color and render each character in T-pose with different perspectives to create the color design sheets.
We also utilize toon shading techniques to render the shadows and highlights to create shading annotations and color lines for target frames.
Based on these newly added annotations, our dataset meets the same standard as the real hand-drawn animation in the animation industry.

This work builds upon the earlier conference version~\cite{InclusionMatching2024} that was accepted by CVPR 2024. Compared with the conference version, we have introduced a significant amount of new materials as follows:
1) Based on the industrial need for hand-drawn animation colorization, we first define color-design-sheet-based paint bucket colorization and divide it into two tasks: \textit{keyframe colorization} and \textit{consecutive frame colorization} (as presented in our earlier conference version~\cite{InclusionMatching2024}).
2) Compared with the previous \pbc~dataset, we provide more annotations including shading annotations and a color design sheet for each character.
3) To achieve accurate keyframe colorization, we propose a two-stage method, consisting of segment parsing and inclusion matching.
4) Extensive experiments illustrate that our keyframe colorization method can support different formats of input, including semantic strokes to help digital painters colorize in a faster and more flexible way.
5) We present a further comprehensive analysis for consecutive frame colorization, including visual comparisons for non-character scene colorization, accuracy assessments across segments of varying sizes, and detailed visual comparisons of colorization methods utilizing ControlNet~\cite{controlnet}.
}

\vspace{-1mm}
\section{Animation Production Process}
\pami{
To clarify the actual need for hand-drawn animation colorization, we revisit the animation production process in this section.
Figure~\ref{fig:ani_production} provides an overview of the basic hand-drawn animation production workflow. 
(a) The process begins with character and color designers creating a character color design sheet, which serves as a reference for the entire animation.
(b) Key animators then draft keyframes, also called layouts, based on the character design sheet.
(c) Animators, also known as inbetween animators, then clean up these keyframes for seamless connection and complete the intermediate frames.
(d) Digital painters, following the color design sheet, colorize each line-enclosed segment of the frames.
(e) Finally, special effect artists tone the colorized frame and adds special effect such as the glow effect shown in Fig.~\ref{fig:ani_production} for specific regions marked with assigned colors.

As stated above, each frame's completion involves multiple staff members. To mitigate potential risks of confusion associated with transferring the project file, explicit guidelines are established for each process.
During the clean-up phase in Fig.~\ref{fig:ani_production}(c), animators use color lines to mark the lines that also need to be colorized.
To assist digital painters in distinguishing between shadows and highlights, shadow areas are enclosed with blue lines, while highlight regions are surrounded by red lines.
However, color lines alone are insufficient to denote shadows or highlights.
Therefore, animators are required to use bright colors to fill in these shadows and highlights. 
Typically, highlights are colored in yellow, while shadows are generally filled with light blue. 
Additionally, animators also need to use different colors to distinguish various parts, such as hair, skin, and others. For instance, as shown in Fig.~\ref{fig:ani_production}, shadows on the skin are marked in light red to differentiate them from those on the hair~\cite{kyoto_draw}.
These annotations are referred to as shading annotations. By adhering to the shading annotations and the color design sheet, digital painters can colorize the line art without confusion.

In this paper, we follow the standard animation production process to synthesize our \pbc~dataset.
Our dataset includes several components that are rarely found in general datasets, such as color lines, shading annotations, and color design sheets.
Moreover, we designed specialized methods to accurately colorize line arts based on these annotations.
Detailed data synthesis process and our methods are provided in Section~\ref{sec:dataset} and \ref{sec:method}.
}

%% file: sec/2_related_work.tex
\begin{figure*}[h]
  \centering
    \vspace{-2mm}
   \includegraphics[width=0.9\linewidth]{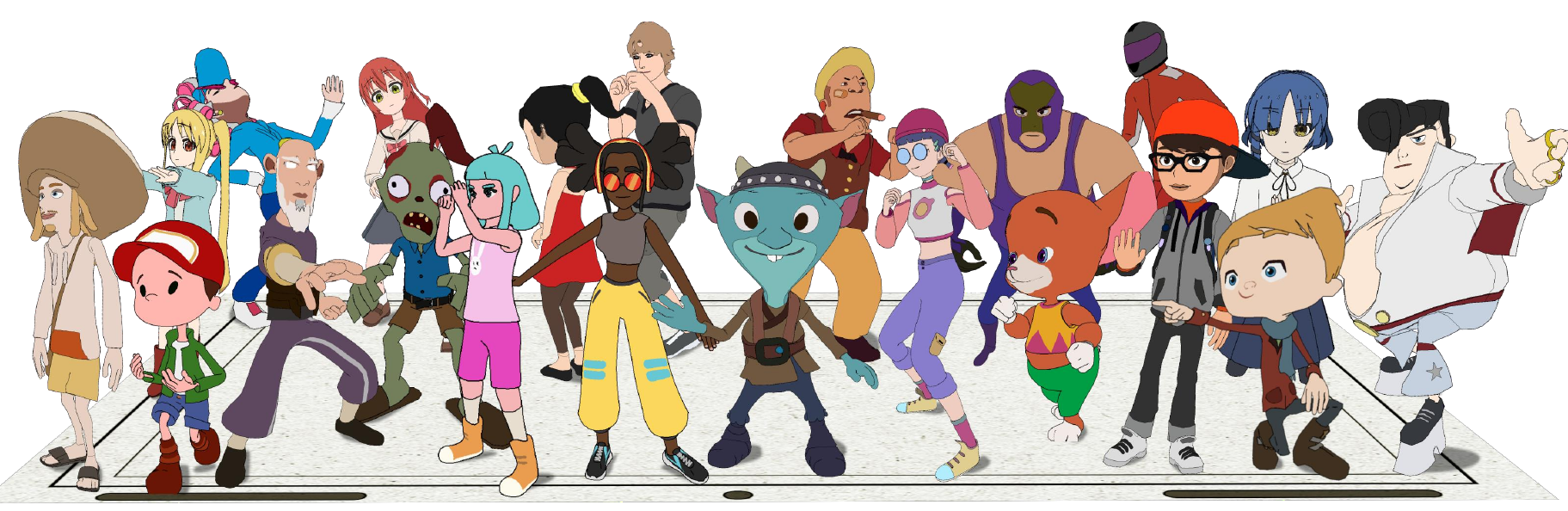}
   \vspace{-3mm}
   \caption{Several examples of our rendered characters in \pbc~dataset.}
   \vspace{-6mm}
   \label{fig:dataset}
\end{figure*}
\section{Related Work}
\label{sec:formatting}

\vspace{2mm}
\noindent{\bf Line Art Colorization.}
Line art colorization aims to create reasonable colors in the sketch's blank regions. To achieve precise control, many user guidance types are designed including text~\cite{kim_tag2pix_2019,zouSA2019sketchcolorization}, scribble~\cite{sykora2009lazybrush, xia-2018-invertible, Filling2021zhang, cao2021line}, colorized reference~\cite{sykora2004unsupervised,sykora2009rigid,chen2020active, lee2020reference,wu2023self,zhang2021line,controlnet,akita2023hand},\pami{\cite{cao2023animediffusion,xing2024tooncrafter,huang2024lvcd}}. 
These methods mainly focus on pixel-wise estimation and cannot fill flat accurate RGB values assigned by the color designer in line arts' segments, which always leads to flickering and color bleeding. 
Besides, in the post-processing stage, the special effect artists often need to select specific colors to create various effects such as ramp or glow for the character's specific region. Thus, current line art colorization methods are still difficult to merge into animation industrial processes.

\vspace{2mm}
\noindent{\bf Animation-related Datasets.} To facilitate hand-drawn animation production, many real and rendered datasets are proposed. 
To boost video interpolation performance, Li~\etal~\cite{siyao_deep_2021} propose the ATD-12K dataset which consists of triplet frames from 30 real animation movies. 
\pami{Xing~\etal~\cite{xing2024tooncrafter} collect a cartoon video dataset with 271K cartoon video clips for anime generative interpolation.}
Due to the absence of annotations for real animation, 3D rendering techniques have been widely used to render these necessary annotations.
By using Blender, MPI-Sintel~\cite{sintel} and CreativeFlow+~\cite{shugrina2019creative} can provide different annotations such as optical flow, segmentation labels, and depth maps for multiple applications. 
To overcome the domain gap between rendered images and real 2D cartoons, Li~\etal~\cite{siyao2022animerun} develop the AnimeRun dataset using open-source 3D animations and their rendered optical flow. 
It improves the accuracy of segment matching and handles complex scenes.

\vspace{2mm}
\noindent{\bf Segment-based Animation Colorization.}
In the cleaning-up process of the animation industry, animators enclose each line manually, making segments serve as basic units for colorization. Thus, calculating segment-level semantic information and correspondence becomes the key to animation colorization. Traditional methods~\cite{maejima2019graph,liu_shape_2023,liu2020shape,zhu-2016-toontrack,zhang_excol_2012} regards segment as node and adjacency relationship as edge, which transfer the segment matching to the graph optimization problem. 
Recently, Dang~\etal~\cite{dang_correspondence_2020} use Hu moments~\cite{hu1962visual} to extract each segment's feature to replace the RGB value and apply a UNet structure~\cite{unet} to obtain the feature map. 
Subsequently, the feature is averaged in each segment to calculate the distance across frames for matching. 
Referring to the success of Transformer~\cite{vaswani2017attention} in image matching~\cite{sarlin20superglue}, Casey~\etal~\cite{AnT} apply multiplex transformer to aggregate information across frames to obtain more accurate results.
Besides, Casey~\etal replace the Hu moments algorithm with a segment feature extraction CNN which crops and resizes each segment to fixed resolution first.
\pami{Maejima~\etal~\cite{maejima2024continual} propose a new patch-based learning method for few-shot paint bucket colorization, addressing the challenge of limited training data by employing a fine-tuning approach.
Since the code and datasets of Dang~\etal, Casey~\etal, and Maejima~\etal are not publicly available, AnimeRun~\cite{siyao2022animerun} remains the sole benchmark for segment matching available to researchers.}

\pami{

\noindent{\bf Contrastive Learning.}
Contrastive learning aims to learn data representations where similar instances are close in the representation space, while dissimilar instances are far apart.
SimCLR~\cite{chen2020simclr} introduces a contrastive learning framework of visual representations by maximizing the similarity between differently augmented views of the same image using a contrastive loss in the latent space.
Moco~\cite{he2020moco} presents a framework for unsupervised visual representation learning that operates as a dynamic dictionary look-up system.
Recently, contrastive learning has been widely used in multimodal learning to align different modalities.
CLIP~\cite{clip} and ALIGN~\cite{jia2021scaling} use contrastive learning to train text and image encoders that align the two modalities.
In this paper, we build on the approach of \cite{clip}\cite{jia2021scaling} which applies contrastive learning to align the text and image to match the target frame's image feature with the reference frame's text description.
}

%% file: sec/3_dataset.tex
\begin{figure}[t]
  \centering
   \includegraphics[width=1.0\linewidth]{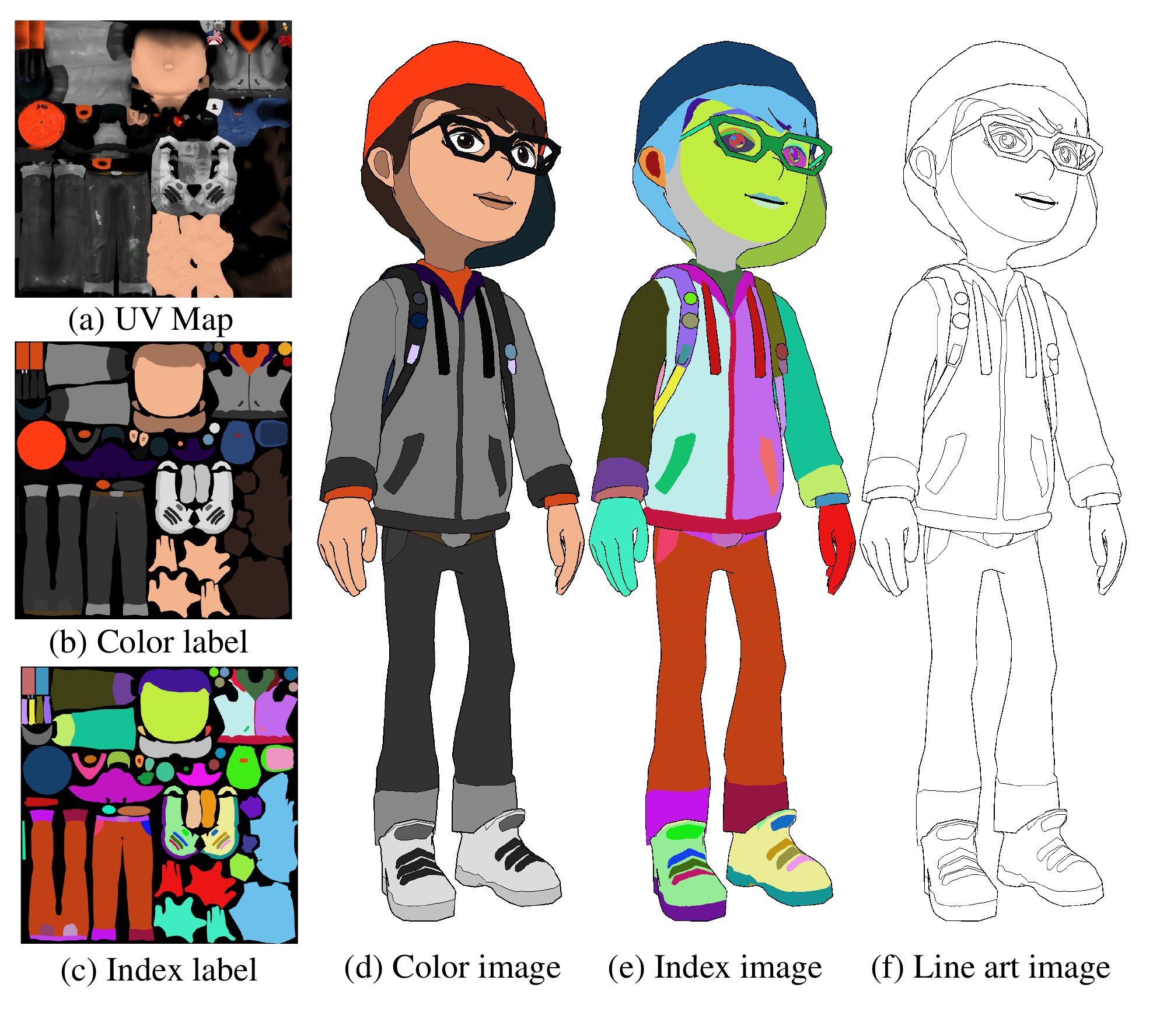}
   \vspace{-8mm}
   \caption{Overview of our synthetic data generation pipeline. We extract the UV map (a) from the character. Then, we use a paint bucket tool to fill the occupied UV region, creating color label (b) and index label (c). Label images (b) and (c) are then pasted back to the 3D meshes to create flat color characters (d) and (e), respectively. Finally, we post-process the index image (e) to obtain the rendered line art image (f).}
   \vspace{-5mm}
   \label{fig:data_generation_pipeline}
\end{figure}

\begin{figure*}[t]
  \includegraphics[width=\textwidth]{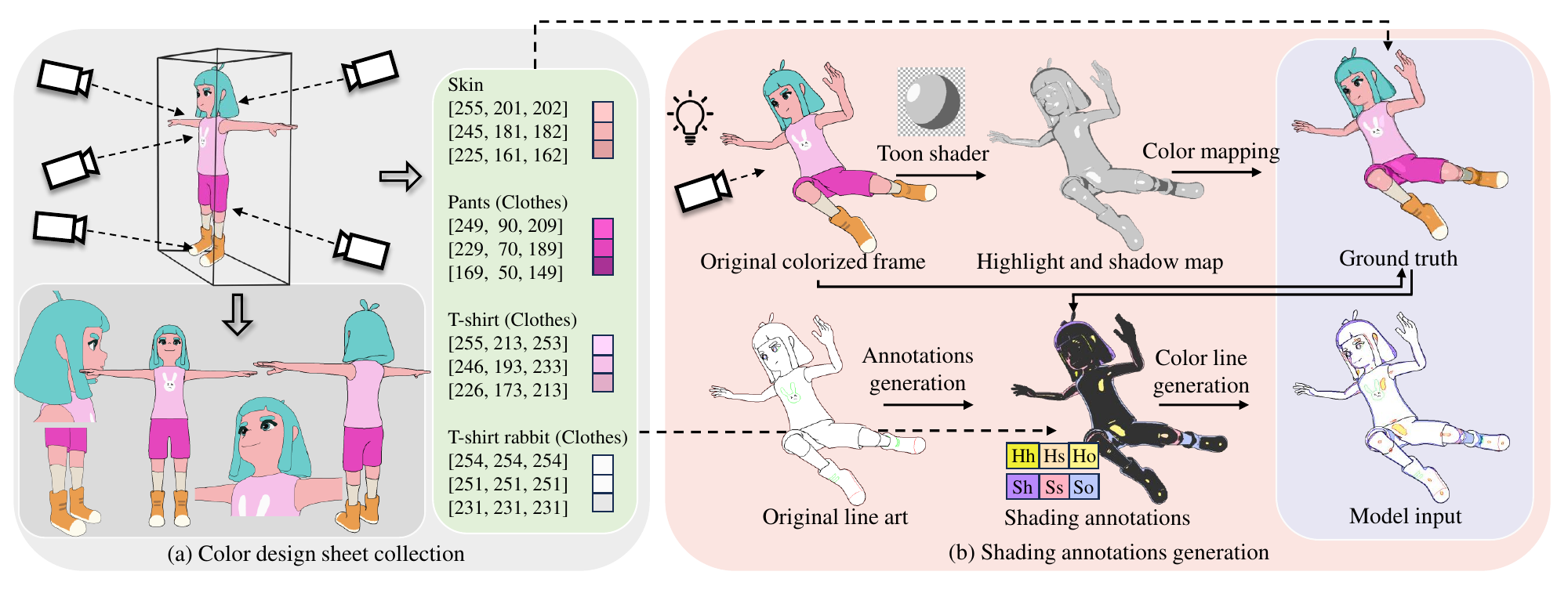}
  \vspace{-8mm}
  \caption{\pami{Our data generation pipeline for (a) the color design sheet and (b) ground truth and model input with shading annotations. In particular, we use `Hh, Hs, Ho, Sh, Ss, So' as shading annotations to indicate the highlight and shadow regions of hair, skin, and others.}}
  \label{fig:data_generation}
  \vspace{-6mm}
\end{figure*}
\begin{figure}[!t]
  \centering
   \includegraphics[width=1.0\linewidth]{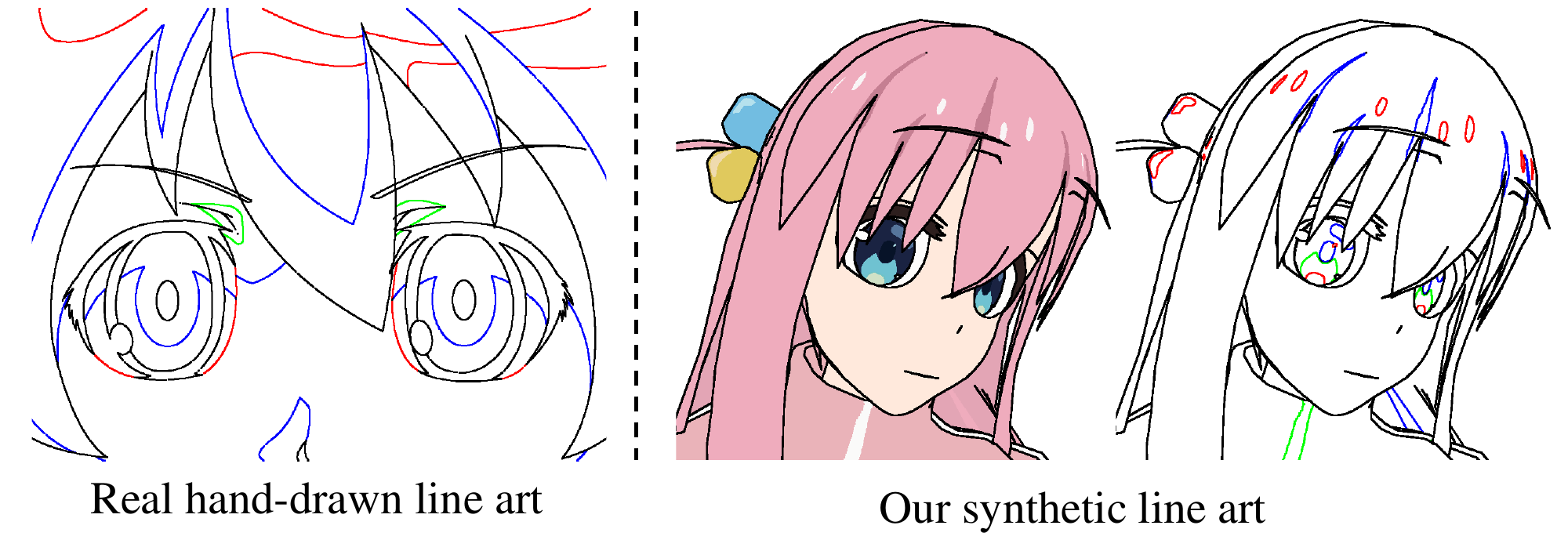}
   \vspace{-8mm}
   \caption{Our data synthesis method generates color lines that closely resemble hand-drawn line art. Red, blue, and green 
 lines represent highlights, shadows, and other instructions, respectively.}
   \label{fig:line_generation_pipeline}
\end{figure}

\section{Paint Bucket Colorization Dataset} \label{sec:dataset}
We introduce a new dataset~\pbc, comprising 11,345 training images and 3,200 test images. The test set includes 3,000 3D-rendered frames and 200 hand-drawn frames. 
Notably, our dataset focuses solely on character animations, considering the common practice among animators of drawing foreground character animations while employing 3D models or separate paintings for the background.
To build our dataset, we sample 22 characters from Mixamo~\cite{mixamo} and Aplaybox~\cite{aplaybox}, including 12 characters for training and 10 for testing.
As depicted in Fig.~\ref{fig:dataset}, our dataset showcases a diverse array of characters, encompassing both Japanese and Western cartoon styles. 

To offer rich temporal and occlusion relationships in our dataset, for each Mixamo's character, we add five types of actions for each character and move the camera to capture different viewpoints of the character in each sequence. We categorize the camera positions into three types: face shot, long shot, and close-up, focusing on the face, entire body, and specific body parts. 
Using this approach, we generate 15 clips per character, yielding 180 clips for training and 150 clips for testing.
In animation, keyframes typically have a low frame rate~\cite{siyao_deep_2021}. To simulate this, we set the frame rate to 10 in each clip, bringing larger motion and deformation. 
We set our test image size to 1024$\times$1024 to ensure that most algorithms can be evaluated on our benchmark.
Meanwhile, to capture fine details such as pupils and buttons, we use a training image size of 1920$\times$1920 to provide clear segments in these areas.

Our proposed dataset is carefully designed with the following key properties: a) a flat color style that adheres to anime production standards, b) accurate index labels and precise line art to define each color segment, c) color design sheets with semantic categories, and d) shading annotations, including shadow and highlight regions, to distinguish confusing areas and to facilitate the task of keyframe colorization. 
The major steps involved in constructing our dataset are detailed as follows.

\pami{
\subsection{Synthetic Dataset Construction}
\noindent{\bf Step-1: Flat Color Rendering.} 
The process is depicted in Fig.~\ref{fig:data_generation_pipeline}.  
To render characters in a flat color style, which is common in anime production, we first use Cinema 4D to extract a UV mesh map and set unused regions as black. We then draw contour lines of character details, such as pockets and shoelaces, directly on the UV map. Based on this UV map, we apply a paint bucket tool to colorize the UV regions, creating the color label, shown in Fig.~\ref{fig:data_generation_pipeline}(b). At this step, we also define a text-color list for each character to capture the color of each region, together with its associated shadow and highlight colors. A character with a flat color style can be obtained by pasting the color label back to the 3D meshes, shown in Fig.~\ref{fig:data_generation_pipeline}(d). 

\noindent{\bf Step-2: Define Segments by Index Labels and Line Art.}
We wish to assign a unique index label to a segment with coherent semantics as this can offer a more precise matching relationship for training. To achieve this, we connect color labels with the same semantic meaning (\eg~the palm and back of the hand) and use the connected components algorithm to extract the index label, as shown in Fig.~\ref{fig:data_generation_pipeline}(c). 
A segment can be more clearly defined by its enclosing line art. To generate contour lines for each segment, we calculate the difference between each color segment and its eroded version, producing the black contour lines as shown in Fig.~\ref{fig:data_generation_pipeline}(f). 
This process ensures that each line-enclosed segment consists of discrete, distinct colors, allowing us to easily assign specific meanings to each segment in the subsequent steps.
}

\begin{table*}[]
\centering
\caption{A comparison between our \pbc~dataset and previous segment matching datasets. Since the AnT~\cite{AnT} dataset is not publicly available, we use `-' to mark the unknown information. AnT possesses a private hand-drawn dataset with 3578 frames; however, it does not provide train-test-split information, indicated by `+'.  Seg. per Fr. and Sha. anno. indicate segments per frame and shading annotations.}\label{tab:data_compare} 
\vspace{-3mm}
\resizebox{1.0\textwidth}{!}{
\begin{tabular}{lcccccccccc}

\toprule
\multicolumn{1}{l}{ \multirow{2}*{Dataset} }& \multicolumn{5}{c}{Statistics} &\multicolumn{5}{c}{Annotations}\\
\cmidrule(r){2-6}
\cmidrule(r){7-11}
\multicolumn{1}{l}{}&  Train & Test & Clips & Seg. per Fr. & No anti-aliasing & Optical flow & Index label & Color line & \pami{Sha. anno.}  & \pami{Color design sheet}\\
\toprule
AnimeRun~\cite{siyao2022animerun}             &  1,760& 1,059& 20+10& 237    &   $\times$&$\checkmark$&$\times$ & $\times$&\pami{$\times$}&\pami{$\times$}\\
Creative Flow+~\cite{shugrina2019creative} &  124k & 10k & 2490+810 & 49 & $\times$ & $\checkmark$ & $\times$ & $\times$ & \pami{$\times$} & \pami{$\times$} \\
AnT~\cite{AnT}~(private)  &  9,900+& 1,100+& -& $<$ 50      &   -&$\times$&$\checkmark$& $\times$ & \pami{$\times$}& \pami{$\times$}\\
Ours                   &  11,345& 3,200& 180+170& 169     &   $\checkmark$&$\times$& $\checkmark$ & $\checkmark$ & \pami{$\checkmark$} &\pami{$\checkmark$}\\
\hline
\end{tabular}
}
\vspace{-3mm}
\end{table*}

\begin{figure*}[t]
  \centering
   \includegraphics[width=1.0\linewidth]{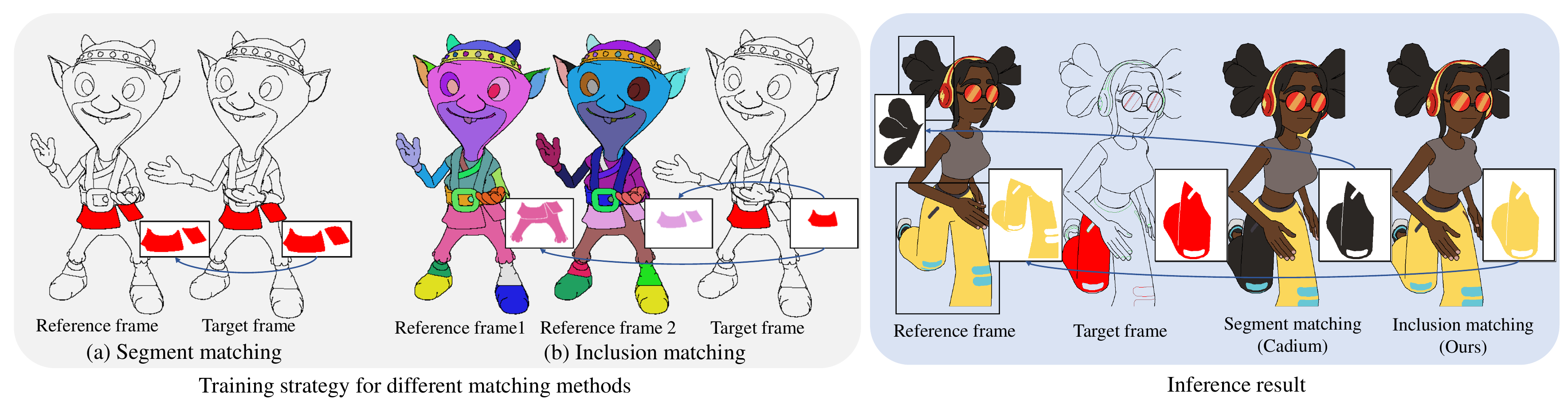}
   \vspace{-8mm}
   \caption{\pami{Training strategy of inclusion matching. Unlike previous segment matching methods~\cite{AnT,siyao2022animerun}, inclusion matching method learns to match each line-enclosed segment in the target frame with a corresponding index label from the reference frame, instead of relying on direct label-to-label matching.
   As demonstrated in our inference results, this training strategy can guide the network to learn the inclusion relationships, enabling it to match, for instance, the red segment in the target frame with the masked yellow segment in the reference frame, even in cases where shape similarity is lacking.}
   }
   \vspace{-5mm}
   \label{fig:inclusion_matching}
\end{figure*}

\pami{
\noindent{\bf Step-3: Color Design Sheets with Semantic Categories.}
In this work, we address a new task called keyframe colorization, which focuses on colorizing keyframes based on a provided character color design sheet.
A color design sheet consists of several images showing different views of a standard pose of the character and a predefined text-color list, known as color setting, as shown in Fig.~\ref{fig:data_generation}(a). 
The color design sheet is commonly used to instruct digital painters on colorization and inform them of the highlight and shadow color values for each semantic element.
For instance, the highlight of the skin region should be colored with \text{``[255, 201, 202]''} in the example.

We generate a color design sheet for each character in our synthetic dataset. We first set the character in a T-pose and alter the camera viewpoint to capture a close-up, face shot, and long shot as reference images in the color design sheet. The text-color list is obtained in Step-1 when we colorize the UV map.
To facilitate the training process, we manually classify the text descriptions into 12 major semantic categories commonly found on an anime character, including background, bag, belt, glasses, hair, socks, hat, mouth, clothes, eyes, shoes, and skin.

\noindent{\bf Step-4: Shading Annotations.}
In animation production, animators use annotations to label the highlight and shadow regions.
To generate these regions, we first apply a toon shader, a non-photorealistic rendering (NPR) technique, to generate the character's highlight and shadow map. We disable anti-aliasing and use nearest-neighbor sampling for the shader. This process is illustrated in Fig.~\ref{fig:data_generation}(b). 
The final colorized ground truth is created by referencing the corresponding highlight and shadow colors from the text-color list, e.g., skin and clothes (the color setting shown in Fig.~\ref{fig:data_generation}(a)), and applying them to the original rendered colorized frame based on the highlight and shadow map.
%
To synthesize the model input based on the original line art, we apply the same operation described in Step-2 to the shadow and highlight regions to extract their contour lines. These lines are shown in red, blue and green colors, as illustrated in Fig.~\ref{fig:line_generation_pipeline}. In addition, we define six labels to distinguish the shadow and highlight regions for hair, skin and others. These labels are referred to as shading annotations (Fig.~\ref{fig:data_generation}(b)).
}

\noindent{\bf Comparisons with Existing Datasets.}
We compare our dataset~\pbc~with previous dataset AnimeRun~\cite{siyao2022animerun}, Creative Flow+~\cite{shugrina2019creative} and AnT~\cite{AnT} in Table~\ref{tab:data_compare}.
The proposed dataset outnumbers both AnimeRun and AnT in terms of training and test images. The additional training images help mitigate the risk of network overfitting. Note that Creative Flow+ contains more data but does not provide the rich annotations we do, including the index labels, color lines, shading annotations, and color design sheets widely used in real animation production. 
The color design sheets and shading annotations make our dataset unique for training and evaluating keyframe colorization.
It is noteworthy that AnimeRun and Creative Flow+ use rendered optical flow to determine segment correspondence. Such an approach often results in frequent one-to-many and many-to-one correspondences in the ground truth, leading to ambiguity during evaluation. Our approach resolves the problem through index labeling.
In addition, due to the anti-aliasing applied to AnimeRun's and Creative Flow+'s rendered images, accurate color labels cannot be reliably assigned to each segment. Consequently, these datasets face difficulties when used directly as a paint bucket colorization dataset.

\subsection{Hand-drawn Dataset Collection}
Obtaining intermediate unprocessed data is challenging, as most commercial animations undergo post-processing techniques like anti-aliasing. To overcome this, we collaborate with professional animators to create line art animation clips featuring different characters. We then use the paint bucket tool to colorize these line art frames, creating ground truth data for synthesizing paired line art and colorized frames. Besides, animation software like Retas Studio Paintman and CLIP Studio Paint offer animation production tutorials. We collect animations from these tutorials and apply our color line extraction method to synthesize paired data. Lastly, we compile a hand-drawn test dataset consisting of 200 frames across 20 clips for evaluation.

%% file: sec/4_method.tex
\begin{figure*}[ht] 
  \centering
   \includegraphics[width=1.0\linewidth]{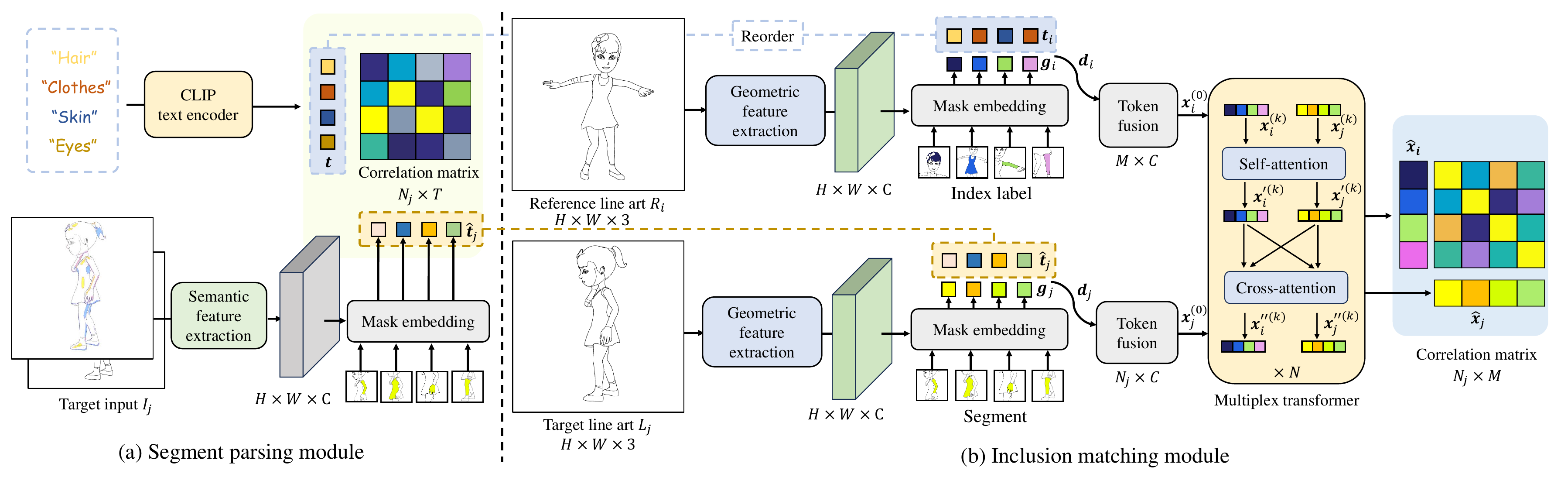}
   \vspace{-6mm}
   \caption{\pami{Architecture of \textit{keyframe colorization.} Initially, we estimate each segment's semantic category based on the line arts and shading annotations using contrastive learning. These extracted semantic tokens are then concatenated with the extracted geometric tokens and fused together using token fusion module, which is then fed into the multiplex transformer for information aggregation. Finally, we compute a similarity matrix between reference and target tokens to calculate matching loss and predict the final color for each segment.} }
   \label{fig:pipeline2}
   \vspace{-4mm}
\end{figure*}
\vspace{2mm}

\section{Methodology} \label{sec:method}
\subsection{Inclusion Matching}
Existing segment-matching-based methods~\cite{AnT,siyao2022animerun} mainly focus on calculating segments' visual correspondence based on segments' shape similarities and relative positions.
Specifically, during the training process,  these methods learn to identify the most similar index label in the reference frame for a given index label in the target frame, as illustrated in Fig.~\ref{fig:inclusion_matching}(a).
It is important to note that there is typically a clear one-to-one correspondence between index labels in the reference and target frames.
However, during the inference stage, line-enclosed segments do not have one-to-one correspondences.
Wrinkles and occlusions may break this correspondence and result in one-to-many and many-to-many correspondence.
This domain gap makes it challenging for segment-matching methods to handle occlusions effectively during inference.

To address this issue, we propose a novel training strategy called \textit{inclusion matching}, which guides the network to understand the inclusion relationship between segments rather than relying solely on direct visual correspondences.
During the inference stage, our method can match target frame's segment with its corresponding larger reference frame's segment that includes it.
As shown in Fig.~\ref{fig:inclusion_matching}(b), the network learns to identify which specific color region in the reference frame includes the red-marked segment in the target frame.
In this case, the network learns that this target segment is included in the pink region of reference frame 1 and the purple region of reference frame 2.

\pami{
Learning the inclusion relationship is non-trivial.
Depending on how segments are grouped, the same target segment may be classified into different segment groups for the same reference line art, leading to varied matching outcomes, as shown in Fig.~\ref{fig:inclusion_matching}(b). 
Therefore, the mechanism of encoding the grouping information of the line art into the network is crucial for accurate inclusion matching.
While the grouping information of the reference frame can naturally be derived from its color, the main challenge lies in estimating a coarse grouping for the target frame.

In Section~\ref{sec: method_key} and Section~\ref{sec: method_consecutive}, we will introduce the overall pipeline for \textit{keyframe colorization} and \textit{consecutive frame colorization}, and discuss how we estimate the target frame's coarse grouping information and encode the grouping information for training.
In \textit{keyframe colorization}, the goal is to colorize a line art with an arbitrary pose based on a given color design sheet.
This sheet consists of multiple reference images and a predefined text-color list.
In \textit{consecutive frame colorization}, we aim to colorize a line art based on a colorized reference frame in a similar pose.
}

\subsection{Keyframe Colorization}\label{sec: method_key}
\pami{
\noindent{\bf Formulation.}
The architecture of \textit{keyframe colorization} is depicted in Fig.~\ref{fig:pipeline2}.
For a color design sheet with $P$ images, $M$ sets of color entries in the text-color list can be categorized into $T$ semantic categories. Each color entry marks the unique normal, highlight, and shadow color.
The reference set is denoted as $D_{ref} = \{(R_i,\mathbf{s}_i)\}_{i=1}^{P}$, where $R_i \in \mathbb{R}^{H \times W \times 3}$ is the line art with $M$ index labels and $\mathbf{s}_i \in \{0,1\}^{M \times T}$ represents the semantic category for each index label in one-hot representation.
In the training stage, each color entry is represented by the index label.
Our task aims to estimate the specific color entry for each line-enclosed segment $ \mathbf{\hat y}_j \in [0,1]^{N_j \times M}$ for a given line art $L_j \in \mathbb{R}^{H \times W \times 3}$ with $N_j$ line-enclosed segments and its corresponding shading annotations.
Since highlight and shadow regions in the shading annotations will not influence the color entry estimation result, we use three parsing masks $S_j \in \mathbb{R}^{H \times W \times 3}$ to mark the regions of the hair, skin, and others to represent the shading annotations. 
It is mainly used to help distinguish some confusing regions in semantics.

\noindent{\bf Overall Pipeline.}
The architecture for \textit{keyframe colorization} consists of a segment parsing module and an inclusion matching module.
The segment parsing module is responsible for estimating the semantic category of each segment in the target frame, facilitating the matching of corresponding segments with the same semantics.
These semantic categories act as the grouping information and are fused with the features of each segment, which are extracted from the line art in the token fusion module of Fig.~\ref{fig:pipeline2}(b).
This design enables inclusion matching based on both semantic and geometric features, leading to more robust performance, as shown in Fig.~\ref{fig:matching_problem}.

In the segment parsing module shown in Fig.~\ref{fig:pipeline2}(a), the line art $L_j$ and shading annotations $S_j$ of the target frame are concatenated to create the target input $I_j$.
Then, a semantic feature extraction module is applied to extract its pixel-wise feature $ \mathbf{f}_j \in \mathbb{R}^{H \times W \times C}$. We apply a mask embedding module to average the feature within each segment's mask, transforming the feature map into a semantic sequence $ \mathbf{\hat t}_j \in \mathbb{R}^{N_j \times C}$.

In the inclusion matching module depicted in Fig.~\ref{fig:pipeline2}(b), the geometric feature extraction module and mask embedding module function tokenize the index labels of the reference line art and segments of the target line arts to geometric sequences $\mathbf{g}_i \in \mathbb{R}^{M \times C}$, $\mathbf{g}_j \in \mathbb{R}^{N_j \times C}$ in the same way as the segment parsing module.
The semantic sequence of the reference frame, denoted as $\mathbf{t}_i \in \mathbb{R}^{M \times C}$, represents the semantic category's feature of each segment. It is obtained by multiplying $\mathbf{s}_i$ with the text embeddings $\mathbf{t} \in \mathbb{R}^{T \times C}$ corresponding to the semantic categories.
The token fusion module then concatenates and fuses the tokens of geometric sequences $\mathbf{g}_i$, $\mathbf{g}_j$ with the semantic sequences $\mathbf{t}_i$, $\mathbf{\hat t}_j$.
Positional embedding is also applied in the token fusion module. 
Then, we apply $N$ transformers for feature aggregation and use the final output of transformers $\mathbf{\hat x}_i\in \mathbb{R}^{M\times C}$ and $\mathbf{\hat x}_j\in \mathbb{R}^{N_j\times C}$ for similarity matrix calculation, yielding the index label prediction of each segment $\mathbf{\hat y}_j$.

\begin{figure}[t]
  \centering
  \includegraphics[width=\linewidth]{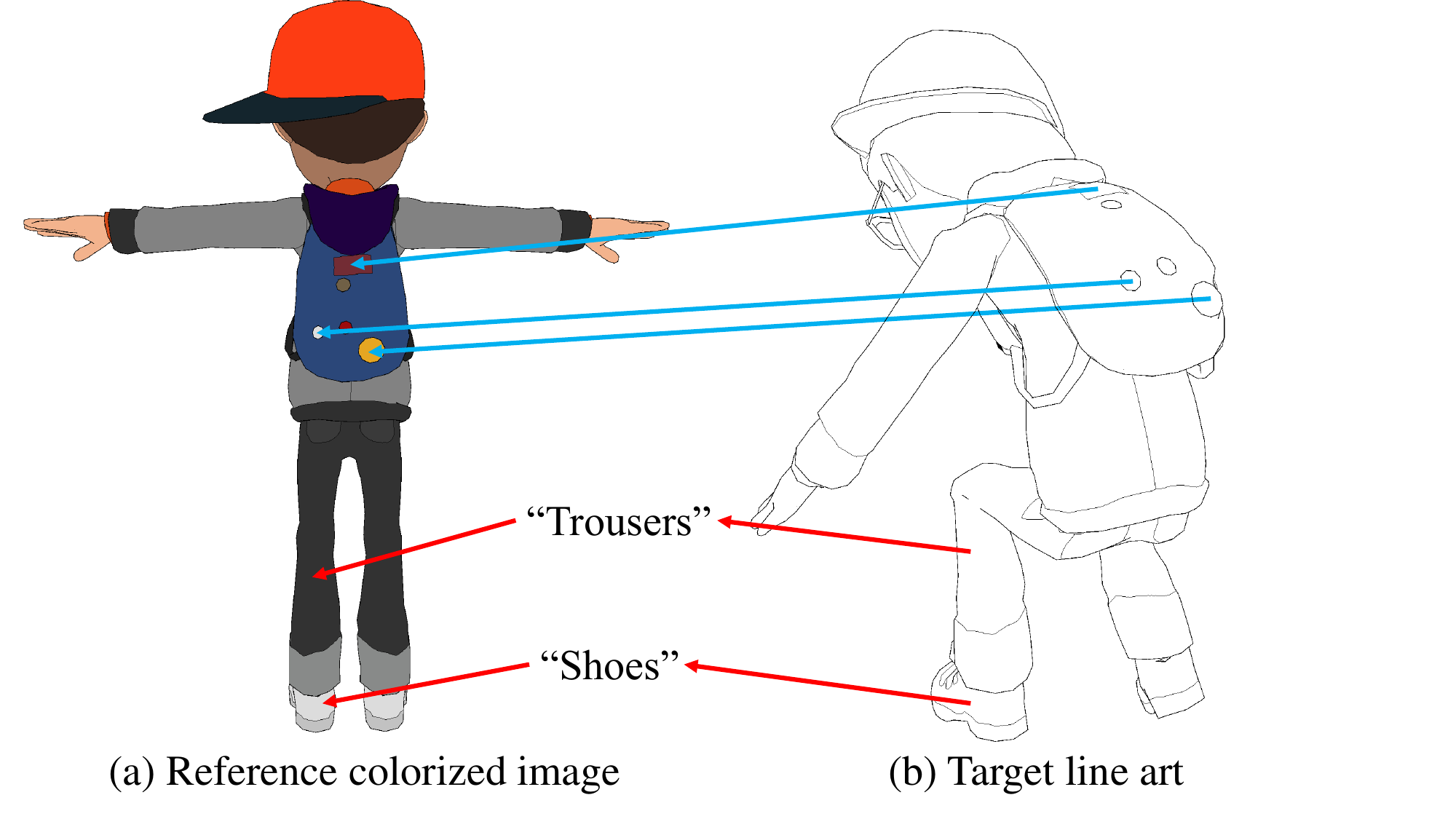}
  \vspace{-3mm}
  \caption{\pami{Semantic features help match corresponding regions within the same semantic category (\eg~red matching lines), while geometric features align segments with similar shapes or spatial relationships (\eg~blue matching lines). Our token fusion module combines these two types of features, enabling effective region matching even when clear shape correspondences or explicit semantic cues are absent.}
  }
  \label{fig:matching_problem}
  \vspace{-3mm}
\end{figure}

\begin{figure*}[h]
  \centering
   \includegraphics[width=1.0\linewidth]{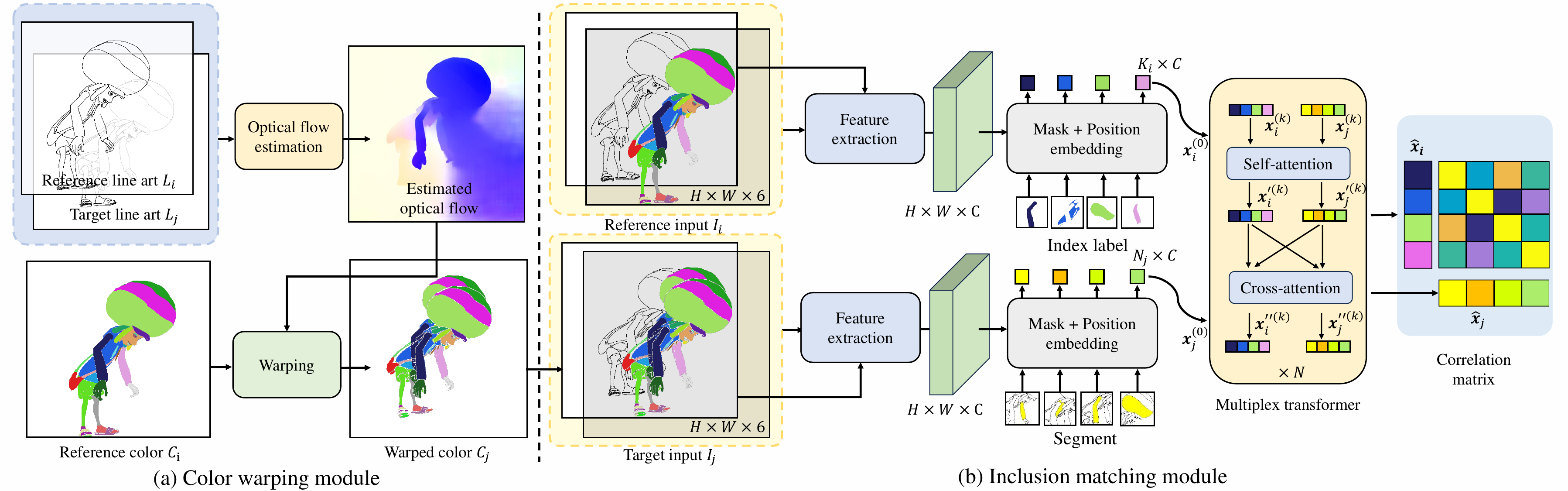}
   \vspace{-6mm}
   \caption{\pami{Architecture of \textit{consecutive frame colorization}}. First, we estimate the optical flow based on the line art and warp the reference colors to generate a coarse colorization result. The feature extraction module then processes the concatenated line art and coarse color to extract relevant features. Next, the mask and position embedding module tokenizes these features into a sequence, which is passed to the multiplex transformer for information aggregation. Finally, a similarity matrix is computed between the reference and target tokens to calculate the matching loss and predict the final color for each segment.}
   \vspace{-5mm}
   \label{fig:pipeline}
\end{figure*}

\noindent{\bf Feature Extraction Network.}
Following \cite{AnT}, we apply a feature extraction network to obtain discriminative features for each segment.
The semantic feature extraction network in Fig.~\ref{fig:pipeline2}(a) and geometric feature extraction network in Fig.~\ref{fig:pipeline2}(b) share the same feature extraction network structure but differ in their weights and number of input channels.
The semantic feature extraction network takes $I_j \in \mathbb{R}^{H \times W \times 6}$ as input, while the geometric feature extraction network takes a 3-channel line art as input. Both networks output a feature map in $\mathbb{R}^{H \times W \times C}$.
Inspired by the FILM network structure~\cite{reda2022film}, the networks adopt a U-Net architecture with input pyramids. These input pyramids are designed to mitigate the impact of varying line thickness. 
Since line art at high resolution contains large blank areas, resulting in sparse useful features, we incorporate a Pyramid Pooling Module (PPM)~\cite{zhao2017pyramid, zhou2022lednet} at the network's bottleneck to enhance semantic feature extraction.

\noindent{\bf Mask Embedding.}
To transform the segments into tokens, we employ super-pixel pooling to average the features $\mathbf{f} \in \mathbb{R}^{H \times W \times C}$ extracted by the feature extraction module within each segment's mask, resulting in a $C$-dimensional visual token.

\noindent{\bf Token Fusion.}
Token fusion module is proposed to fuse the semantic sequence $\mathbf{t}_i$, $\mathbf{\hat t}_j$ from Fig.~\ref{fig:pipeline2}(a) with the geometric sequence $\mathbf{\hat g}_i$, $\mathbf{\hat g}_j$ from Fig.~\ref{fig:pipeline2}(b).
Its goal is to combine both semantic and geometric features to facilitate more accurate inclusion matching, as illustrated in Fig.~\ref{fig:matching_problem}.
In the token fusion module, $\mathbf{t}_i$ and $\mathbf{\hat t}_j$ are concatenated with the geometric sequence $\mathbf{\hat g}_i$, $\mathbf{\hat g}_j$ to form the segment's visual tokens $\mathbf{d}_i \in \mathbb{R}^{N_j \times 2C}$ and $\mathbf{d}_j \in \mathbb{R}^{M \times 2C}$.
Since $\mathbf{t}_i$ and $\mathbf{\hat t}_j$ come from different modalities, two distinct Multilayer Perceptron (MLP) layers are applied to $\mathbf{d}_i$, $\mathbf{d}_j$ to reduce their dimensions to $C$.
Subsequently, another MLP layer encodes the segment's bounding box's four-dimensional coordinates, denoted as $\mathbf{\mathbf{p}}_i$, into a $C$-dimensional positional embedding. The token $\mathbf{x}^{(0)}_i$ for matching is obtained by adding the positional embedding and the visual token:
\begin{equation}
\mathbf{x}^{(0)}_{i} = \textbf{MLP}_{i}(\mathbf{d}_{i}) + \textbf{MLP}(\mathbf{\mathbf{p}}_{i}).
\end{equation}

\noindent{\bf Multiplex Transformer.}
Inspired by Casey \etal~\cite{AnT} and Sarlin \etal~\cite{sarlin20superglue}, we employ multiplex self- and cross-attention layers to aggregate tokenized features. In the self-attention layer, queries, keys, and values are derived from a single source feature:
\begin{equation}
\textbf{self-attention}(\mathbf{x}_i) = \text{softmax}\left(\frac{\mathbf{Q}_i\mathbf{K}_i}{\sqrt{D}} \right)\mathbf{V}_i,
\end{equation}
where $\mathbf{Q}_i$, $\mathbf{K}_i$, and $\mathbf{V}_i$ represent $\mathbf{x}_i$ processed by MLPs for query, key, and value. In contrast, the cross-attention layer computes keys and values from another feature:
\begin{equation}
\textbf{cross-attention}(\mathbf{x'}_i,\mathbf{x'}_j) = \text{softmax}\left(\frac{\mathbf{Q'}_i\mathbf{K'}_j}{\sqrt{D}} \right)\mathbf{V'}_j.
\end{equation}
Following each self- or cross-attention layer, the output is added to the original $\mathbf{x}_i$ or $\mathbf{x'}_i$, and then processed with a feed-forward MLP. After repeating these operations $N$ times, we obtain the final aggregated features $\mathbf{\hat x}_i\in \mathbb{R}^{M\times C}$ and $\mathbf{\hat x}_j\in \mathbb{R}^{N_j\times C}$. 

\noindent{\bf Loss Function.}
During the training process, the parsing loss $L_p$ and matching loss $L_m$ are used to optimize the network.
The parsing loss $L_p$ ensures that semantic information is effectively extracted from the segment parsing module, while the matching loss $L_m$ helps achieve accurate matching of both geometric and semantic tokens.
Specifically, in the segment parsing module, we use contrastive learning pipeline to maximize the cosine similarity between the $ \mathbf{\hat t}_j$ and text embeddings $\mathbf{t} \in \mathbb{R}^{T \times C}$ of the text categories using the cross-entropy loss $L_p$.
In the inclusion matching module, the final output $\mathbf{\hat x}_i$ and $\mathbf{\hat x}_j$ are used to calculate the matching loss $L_m$ in the same way.
In these modules, the corresponding similarity matrices are represented as $\mathbf{\hat S}^P \in [0,1]^{N_j \times T}$ and $\mathbf{\hat S}^M \in [0,1]^{N_j \times M}$, respectively. These matrices are derived as follows:
\begin{equation}
{\hat S}^P_{mn} = \frac{\exp(\mathbf{t}_{im} \cdot \mathbf{\hat t}_{jn})}{\sum_{m=1}^{N_j} \exp(\mathbf{t}_{im} \cdot \mathbf{\hat t}_{jn})},
\end{equation}

\begin{equation}
{\hat S}^M_{mn} = \frac{\exp(\mathbf{\hat x}_{im} \cdot \mathbf{\hat x}_{jn})}{\sum_{m=1}^{N_j} \exp(\mathbf{\hat x}_{im} \cdot \mathbf{\hat x}_{jn})}.
\end{equation}

These matrices can be interpreted as the estimated semantic category probability $\hat {\mathbf{p}}_m \in [0,1]^{T}$ and color probability $\hat {\mathbf{y}}_m \in [0,1]^{M}$ for segment $m$ in the target frame. For each frame's one-hot ground truth semantic category $\mathbf{p}_m$ and index label $\mathbf{y}_m$, the loss function is derived from the addition of matching loss and parsing loss:
\begin{equation}
L = -\sum_{m=1}^{N_j} \mathbf{y}_m \log(\hat{\mathbf{y}}_m) -w\sum_{m=1}^{N_j} \mathbf{p}_m \log(\hat{\mathbf{p}}_m),
\end{equation}
where $w$ is set to 0.1 in our experiment.

\noindent{\bf Inference Strategy.} 
During the inference stage, the index labels are not provided. For a reference frame containing $N_i$ line-enclosed segments and $M$ color entries in the text-color list, we primarily estimate a similarity matrix $\mathbf{\hat S}^M \in [0,1]^{N_j \times N_i}$, which represents the probability of each segment in the target frame being included in a segment from the reference frame. Each segment’s color entry can be represented as $ \mathbf{c}_i \in {\{0,1\}}^{N_i \times M}$ in the reference frame.
The estimated color entry for each line-enclosed segment $\hat{\textbf{c}}_j \in [0,1]^{N_j \times M}$ in the target frame is computed as follows:
\begin{equation}
\hat{\textbf{c}}_j = \hat{\textbf{S}}^M \textbf{c}_i.
\end{equation}

Finally, the highlight and shadow areas are colorized by retrieving the corresponding colors from the text-color list, according to the regions specified by the shading annotations.

\noindent{\bf Multi-reference Colorization.} 
A color design sheet always contains multiple reference frames with different perspectives of the character.
When digital painters colorize the line art, they often select just one or two reference images from the reference set with the closest perspective. To mimic this process, we design methods to choose the reference frames that are most similar to the target frame.
Since there are no existing methods to evaluate perspective similarity in line arts, we train an additional network to select the most similar reference frame for a given target line art.
Specifically, we use a pretrained image encoder of CLIP ViT-B/32 model~\cite{clip} to transform the image to a sequence and apply an MLP to extract the perspective information from the sequence.
To train this MLP, we adopt a contrastive learning pipeline. We pick the frame that is close to the target frame in the sequence as the positive sample and another frame that is distant in the sequence as a negative sample. These frames are all preprocessed by the CLIP image encoder, and a triplet loss is applied to optimize the MLP.
In the inference stage, cosine similarities are calculated between the target frame and all other given reference frames.
Then, a softmax layer is applied to calculate the possibility ${p}_k$ for each reference frame $k$.
We only use reference frames with the possibility higher than 0.3 for evaluation.
The estimated color entry for each line-enclosed segment $\hat{\textbf{c}}_j \in \mathbb{R}^{N_j \times M}$ can be derived by:
\begin{equation}
\hat{\textbf{c}}_j = \sum_{\substack{k=1 \\ p_k > 0.3}}^{P} p_k \hat{\textbf{S}}^M \textbf{c}_i.
\end{equation}
If the highest possibility is lower than 0.3, only the first frame in the color design sheet (long shot from the front) will be chosen as the reference frame.

\noindent{\bf Implementation Details.}
Our model undergoes 200,000 iterations with a batch size of 2 on NVIDIA GeForce RTX 3090 GPUs. 
The Adam optimizer with a learning rate of $10^{-4}$ is used without weight decay.
During the training process, the weights of the CLIP text encoder are frozen.
The channel number $C$ for our extracted feature is set to 128.
Within the multiplex transformer, the layer number is set to 9, and the number of heads is configured as 4.
}

\subsection{Consecutive Frame Colorization}\label{sec: method_consecutive}
\noindent{\bf Formulation.} Unlike \textit{keyframe colorization}, where a color design sheet is used, \textit{consecutive frame colorization} only refers a colorized frame with a similar pose.
The architecture of \textit{consecutive frame colorization} is depicted in Fig.~\ref{fig:pipeline}.
Given two adjacent line art images $L_i, L_j \in \mathbb{R}^{H \times W \times 3}$, lines separate the blank area into $N_i$ and $N_j$ line-enclosed segments, respectively.
During training, reference frame $L_i$'s index labels are randomly merged with adjacent index labels with a probability of 20\% as a data augmentation for inclusion matching, which partition the reference frame into $K_i$ regions.
Each merged index label represents a specific color during the training process.
The consecutive frame colorization pipeline aims to estimate the index label $\mathbf{\hat y}_j \in [0,1]^{N_j \times K_i}$ for each segment in $L_j$.

\noindent{\bf Overall Pipeline.}
Colorized line art carries substantial information about how segments are grouped, making it crucial for inclusion matching.
The pose similarity between the reference and target frames makes it easier to estimate the target frame's grouping information by using optical flow rather than segment parsing.
Thus, a color warping module is proposed to replace the segment parsing module in the \textit{keyframe colorization} to warp the grouping information.
This warped information is then concatenated with the line art for training, as illustrated in Fig.~\ref{fig:pipeline}(b).

In the color warping module as shown in Fig.~\ref{fig:pipeline}(a), we first recolorize the index label image. Specifically, we partition the RGB space into $K_i$ cubes, each with a side length of $255\times K_i^{-\frac{1}{3}}$. Subsequently, we select the RGB value at the center of each cube and randomly assign them to index labels as the reference color image $C_i \in \mathbb{R}^{H \times W \times 3}$.
We employ the optical flow estimation model RAFT~\cite{teed2020raft}, fine-tuned on AnimeRun~\cite{siyao2022animerun}, to estimate the optical flow between the reference line art $L_i$ and the target line art $L_j$. 
Utilizing this optical flow, we warp the reference color image $C_i$ to generate the coarse color estimation $C_j \in \mathbb{R}^{H \times W \times 3}$ for the target frame.

In the inclusion matching module, the color images $C_i$, $C_j$ are concatenated with the line art $L_i$, $L_j$ as 6-channel input $I_i,I_j \in \mathbb{R}^{H \times W \times 6}$ of the feature extraction network.
Extracted features in $H \times W \times C$ are then tokenized by the mask and positional embedding module as index label and segment sequence $\mathbf{x}_i^{(0)} \in \mathbb{R}^{K_i \times C}$, $\mathbf{x}_j^{(0)} \in \mathbb{R}^{N_j \times C}$.
Same as keyframe colorization, multiplex transformer aggregates the feature as the final output $\mathbf{\hat x}_i \in \mathbb{R}^{K_i \times C}$, $\mathbf{\hat x}_j \in \mathbb{R}^{N_j \times C}$ for similarity matrix calculation, yielding the index label prediction of each segment $\mathbf{\hat y}_j$.

During inference stage, we adopt a color redistribution strategy to assign randomly generated colors to replace the original colors in the color warping module.
This strategy can avoid the mismatching problem when several predefined colors are similar. Finally, the estimated color for each line-enclosed segment $\hat{\mathbf{c}}_j \in [0,1]^{N_j \times K_i}$ can be derived by:
\begin{equation}
\hat{\textbf{c}}_j = \hat{\textbf{S}} \textbf{c}_i,
\end{equation}
where $\hat{\textbf{S}} \in [0,1]^{N_j \times N_i}$ represents the probability of each line-enclosed segment in the target frame being included in the segment of the reference frame and ${\mathbf{c}}_i \in \{0,1\}^{N_i \times K_i}$ represents the color of each line-enclosed segment in the reference frame.

\begin{figure}[t]
  \centering
   \includegraphics[width=1.0\linewidth]{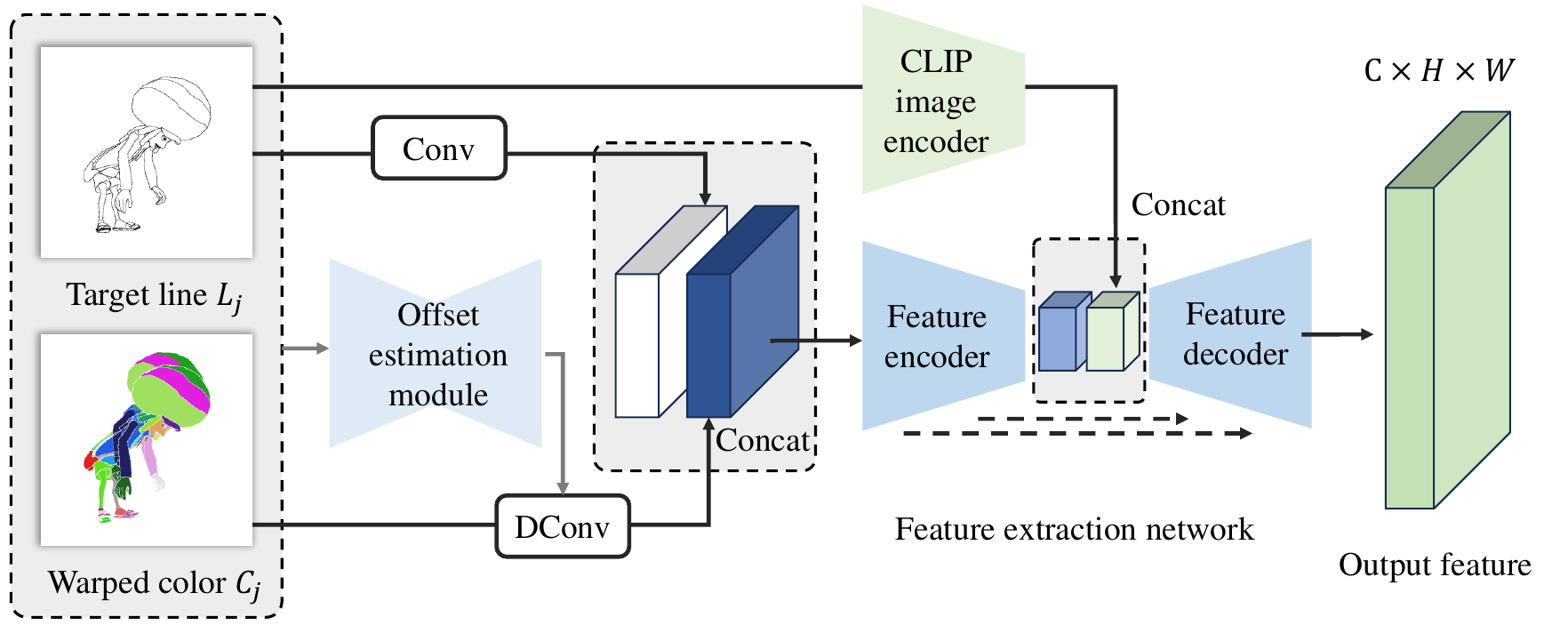}
   \vspace{-7mm}
   \caption{Architecture of our feature extraction module. Deformable convolution is applied to align the color features of the target frame to the line image. Then, the line features and aligned color features are concatenated and fed to the feature extraction network in a U-Net structure. Line art's CLIP features are also concatenated to the bottleneck of the feature extraction network.}
   \vspace{-4mm}
   \label{fig:feature_extraction_network}
\end{figure}

\noindent{\bf Feature Extraction Network.}
Feature extraction network is mainly proposed to extract features from the concatenated line art $L_i$, $L_j$ and color $C_i$, $C_j$.
However, in the target frame, the estimated warped color image $C_j$ cannot precisely align with the line art $L_j$.
To reduce its influence, we apply deformable convolutions~\cite{dai2017deformable} to the warped color $C_j$ to extract aligned features. For a given deformable convolution kernel of $K$ sampling positions, we use $\omega_k$ and $\mathbf{p}_k$ to represent the weight and pre-specified offset. The aligned features $F_{a}$ at each position $\mathbf{p}_0$ can be expressed as:
\begin{equation}
    F_{a}(\mathbf{p}_0)=\sum_{k=1}^{K} \omega_k \cdot F_a(\mathbf{p}_0+\mathbf{p}_k+\Delta \mathbf{p}_k), 
\end{equation}
where learnable offset $\Delta \mathbf{p}_k$ can be predicted from the offset estimation network as depicted in Fig.~\ref{fig:feature_extraction_network}. 
This network denoted as $\Phi$, designed as a lightweight U-Net~\cite{unet}, is employed to expand the receptive field:
\begin{equation}
    \Delta \mathbf{p}_k =\Phi([L_j,C_j]).
\end{equation}
For the reference frame, we deactivate the offset estimation module and set the offset to zero. Subsequently, the aligned color feature is concatenated with the line feature and passed through a feature encoder network to obtain the encoded feature.
To guide the network in learning semantic information matching, we resize the line art image to $320 \times 320$ and utilize CLIP~\cite{clip}'s image encoder to extract features.
To preserve the local contexts better, we choose to use the ConvNext-Large model from OpenCLIP~\cite{cherti2023reproducible}.
We only select the initial layers of CLIP, which downsample the feature to a resolution of $40 \times 40$.
The CLIP feature is interpolated to concatenate with the encoded feature.
Finally, the feature decoder decodes these concatenated features to the input's resolution with $C$ channels. 
To preserve features in high resolution, shortcuts are added between the network's encoder and decoder, following the structure of U-Net.

\noindent{\bf Mask + Position Embedding.}
Same as the mask embedding in \textit{keyframe colorization}, we employ super-pixel pooling to average the features within each segment's mask, resulting in a $C$-dimensional visual token, denoted as $\mathbf{d}_i$.
Subsequently, an MLP layer is utilized to encode the four-dimensional coordinates of the segment's bounding box, denoted as $\mathbf{\mathbf{p}}_i$, into a $C$-dimensional positional embedding.
The token $\mathbf{x}^{(0)}_i$ for matching is obtained by adding the positional embedding and visual token:
\begin{equation}
\mathbf{x}^{(0)}_i = \mathbf{d}_i + \textbf{MLP}(\mathbf{\mathbf{p}}_i).
\end{equation}

\noindent{\bf Multiplex Transformer.}
Our consecutive frame colorization architecture utilizes the same multiplex transformer design as the keyframe colorization architecture.

\noindent{\bf Loss Function.}
After the propagation step, the similarity matrix $\mathbf{\hat S} \in \mathbb{R}^{N_j \times K_i}$ is represented as follows:
\begin{equation}
{\hat S}_{mn} = \frac{\exp(\mathbf{\hat x}_{im} \cdot \mathbf{\hat x}_{jn})}{\sum_{m=1}^{N_j} \exp(\mathbf{\hat x}_{im} \cdot \mathbf{\hat x}_{jn})}.
\end{equation}

This matrix can be interpreted as the color probability $\hat {\mathbf{y}}_m \in \mathbb{R}^{K_i}$ for segment $m$ in the target frame. For each target frame's ground truth index label $\mathbf{y}_m$, the loss function is derived from cross-entropy loss:
\begin{equation}
L_{ce} = -\sum_{m=1}^{N_j} \mathbf{y}_m \log(\hat{\mathbf{y}}_m).
\end{equation}

\begin{table}[]
\caption{ \pami{
Quantitative comparison of our \textit{keyframe colorization} method with different methods. `Acc-Thres' denotes segment-wise accuracy for segments larger than 10 pixels, providing insights into the potential workload reduction for digital painters. `Pix-Acc,' `Pix-F-Acc,' and `Pix-B-MIoU' represent pixel-wise accuracy, foreground pixel-wise accuracy, and pixel-wise background MIoU, respectively, reflecting the visualization performance. For IP-Adapter and LoRA, ControlNet's line art model is used to encode the target frame.}
} \label{tab:comparison_ref}
\resizebox{0.45\textwidth}{!}{
\begin{tabular}{lcccc}
\hline
Method          & \multicolumn{1}{l}{Acc-Thres} & \multicolumn{1}{l}{Pix-Acc} & \multicolumn{1}{l}{Pix-F-Acc} & \multicolumn{1}{l}{Pix-B-MIoU} \\ \hline
IP-Adapter (1-shot)       & 0.1590                        & 0.3772                       & 0.1761                        & 0.4370                          \\
AnimeRun (1-shot)   & 0.2893                        & 0.7495                      & 0.2543                        & 0.8230                         \\
BasicPBC (1-shot)  & 0.5430                        & 0.9012                    & 0.6677                        & 0.9728                         \\
Ours (1-shot)       & 0.6346                          & 0.9531                     & 0.8269                        & 0.9845                          \\ \hline
LoRA (few-shot) & 0.2530                        & 0.6233                      & 0.3667                        & 0.7054                         \\
Ours (few-shot)       & \textbf{0.6459}                          & \textbf{0.9612}                      & \textbf{0.8317}                        & \textbf{0.9867}                          \\ 
\hline
\end{tabular}
}
\end{table}

\begin{figure*}[t]
  \centering
   \includegraphics[width=1.0\linewidth]{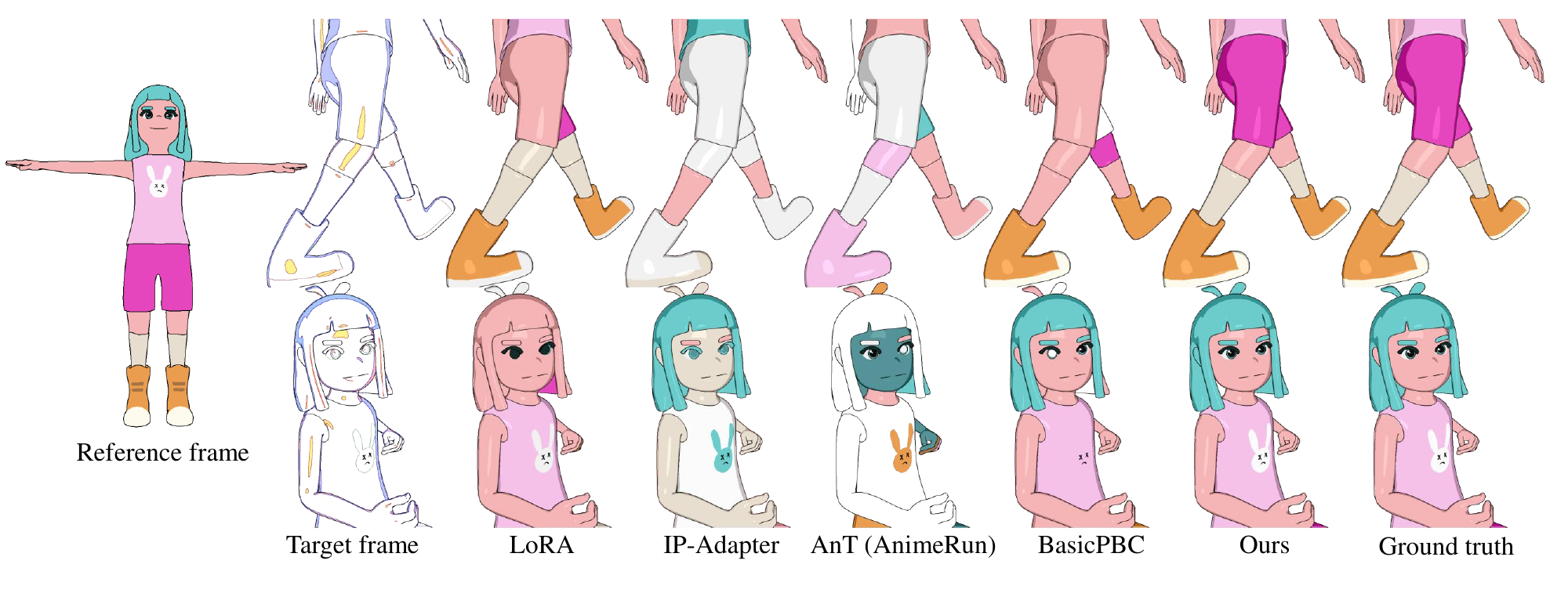}
   \vspace{-12mm}
   \caption{\pami{The visualization comparisons of our proposed \textit{keyframe colorization} methods and other approaches. Our method consistently achieves robust results, even in face shot and close-up scenes.}}
   \vspace{-3mm}
   \label{fig:key_comparison}
\end{figure*}

\begin{table*}[th]
\centering
\caption{Quantitative comparison of our \textit{consecutive frame colorization} method with different methods. `Acc' denotes segment-wise accuracy for all segments. \pami{IP-Adapter means reference-based colorization method using IP-Adapter to encode reference frame and ControlNet to encode target frame.} All RAFT models undergo training on MPI-Sintel~\cite{sintel}, and `F/C' indicates the models fine-tuned on AnimeRun's colorized frames and contours. } \label{tab:baseline}
\resizebox{1.0\textwidth}{!}{
\begin{tabular}{llcccccccccccc}
\hline
                                        \multirow{2}{*}{Type} &         \multirow{2}{*}{Method}                &  & \multicolumn{5}{c}{3D rendered test set}                                                                                                           & \multicolumn{1}{l}{} & \multicolumn{5}{c}{Hand-drawn test set}                                                                                                           \\ \cline{4-8} \cline{10-14} 
                                        &                         &  & \multicolumn{1}{c}{Acc} & \multicolumn{1}{l}{Acc-Thres} & \multicolumn{1}{l}{Pix-Acc} & \multicolumn{1}{l}{Pix-F-Acc} & \multicolumn{1}{l}{Pix-B-MIoU} & \multicolumn{1}{l}{} & \multicolumn{1}{c}{Acc} & \multicolumn{1}{l}{Acc-Thres} & \multicolumn{1}{l}{Pix-Acc} & \multicolumn{1}{l}{Pix-F-Acc} & \multicolumn{1}{l}{Pix-B-MIoU} \\ \cline{1-8} \cline{10-14} 
Reference-based       & IP-Adapter              &  & 0.1709                        & 0.1856                        & 0.4274                        & 0.2653                        & 0.4600                         &                      & 0.1715                        & 0.1812                        & 0.2449                        & 0.2298                        & 0.2260                         \\    
\cline{1-2} \cline{4-8} \cline{10-14} 
\multirow{2}{*}{Segment matching} & AnT(AnimeRun)        &  & 0.5584                  & 0.5930                         & 0.8734                       & 0.6537                        & 0.8954                        &                      & 0.6779                  & 0.7043                         & 0.9193                       & 0.7377                        & 0.9429                        \\
                                        & AnT(Cadmium)                 &  & 0.6634                  & 0.7713                         & 0.9765                       & 0.9471                        & 0.9801                        &                      & 0.7224                  & 0.7879                         & 0.9792                       & 0.9303                        & 0.9899                        \\ \cline{1-2} \cline{4-8} \cline{10-14} 
\multirow{3}{*}{Optical flow}     & RAFT           &  & 0.6589                  & 0.7456                         & 0.9724                       & 0.9099                        & 0.9763                        &                      & 0.7528                  & 0.8093                         & 0.9819                       & 0.9190                        & 0.9898                        \\
                                        & RAFT (F)   &  & 0.7133                  & 0.8054                         & 0.9818                       & 0.9449                        & 0.9852                        &                      & 0.7758                  & 0.8424                         & 0.9860                       & 0.9425                        & 0.9922                        \\
                                        & RAFT (C) &  & 0.7028                  & 0.7944                         & 0.9723                       & 0.9411                        & 0.9690                        &                      & 0.7718                  & 0.8356                         & 0.9833                       & 0.9375                        & 0.9916                        \\ \cline{1-2} \cline{4-8} \cline{10-14} 
Inclusion matching                      & Ours                    &  & \textbf{0.8266}                  & \textbf{0.8726}                         & \textbf{0.9905}                       & \textbf{0.9724}                        & \textbf{0.9948}                        &                      & \textbf{0.8593}                  & \textbf{0.8929}                       & \textbf{0.9900}                       & \textbf{0.9638}                        & \textbf{0.9984}                        \\ \hline
\end{tabular}
}
\vspace{-2mm}
\end{table*}

\begin{table}[]
\caption{ \pami{
Quantitative comparison of the impact of different network structures and methods on \textit{keyframe colorization} test data. `IM' signifies the inclusion matching pipeline, which incorporates adjacent index label merging and segment-label matching strategies. We also retrain AnT~\cite{AnT} and BasicPBC~\cite{InclusionMatching2024} using our dataset for comparison.}
} \label{tab:ablation_ref}
\resizebox{0.45\textwidth}{!}{
\begin{tabular}{lcccc}
\hline
Method          & \multicolumn{1}{l}{Acc-thres} & \multicolumn{1}{l}{Pix-Acc} & \multicolumn{1}{l}{Pix-F-Acc} & \multicolumn{1}{l}{Pix-B-MIoU} \\ \hline
Ours       & \textbf{0.6346}                          & \textbf{0.9531}                     & \textbf{0.8269}                        & \textbf{0.9845}                          \\ \hline
AnT~\cite{AnT}       & 0.5097                          & 0.9013                      & 0.6533                        & 0.9534                          \\
BasicPBC~\cite{InclusionMatching2024}       & 0.5169                          & 0.8916                      & 0.6447                        & 0.9461                          \\ \hline
w/o sha. anno.            & 0.5740                        & 0.9383                        & 0.7523              & 0.9678                        \\
w/o par. loss   & 0.5995                        & 0.9433                      & 0.7695                        & 0.9747                        \\
w/o Inpy  & 0.6047 & 0.9472 & 0.7840 & 0.9781 \\
w/o IM & 0.5974 & 0.9408 & 0.7581 & 0.9751 \\ \hline
\end{tabular}
}
\vspace{-3mm}
\end{table}

\noindent{\bf Inference Strategy.} 
During the inference stage, we adopt a color redistribution strategy to assign randomly generated colors to replace the original colors in the color warping module.
This strategy can avoid the mismatching problem when several predefined colors are similar. Finally, the estimated color for each line-enclosed segment $\hat{\mathbf{c}}_j \in [0,1]^{N_j \times K_i}$ can be derived by:
\begin{equation}
\hat{\textbf{c}}_j = \hat{\textbf{S}} \textbf{c}_i,
\end{equation}
where $\hat{\textbf{S}} \in [0,1]^{N_j \times N_i}$ represents the probability of each line-enclosed segment in the target frame being included in the segment of the reference frame and ${\mathbf{c}}_i \in \{0,1\}^{N_i \times K_i}$ represents the color of each line-enclosed segment in the reference frame.

\noindent{\bf Implementation Details.}
During the training stage, our model undergoes 300,000 iterations with a batch size of 2, utilizing NVIDIA GeForce RTX 3090 GPUs. We employ the Adam optimizer with a learning rate of $10^{-4}$ and no weight decay. Throughout the training process, we freeze the weights of the pretrained optical flow estimation module and the CLIP image encoder.
In the feature extraction module, the offset estimation module is structured as a lightweight U-Net with three down-sampling layers and a bottleneck featuring 128 features. Simultaneously, the feature extraction network is implemented as a U-Net with four down-sampling layers and a bottleneck containing 512 features. The channel number $C$ for our extracted feature is set to 128.
The multiplex transformer is configured with 9 layers and 4 attention heads.

\begin{figure}[t]
  \centering
   \includegraphics[width=1.0\linewidth]{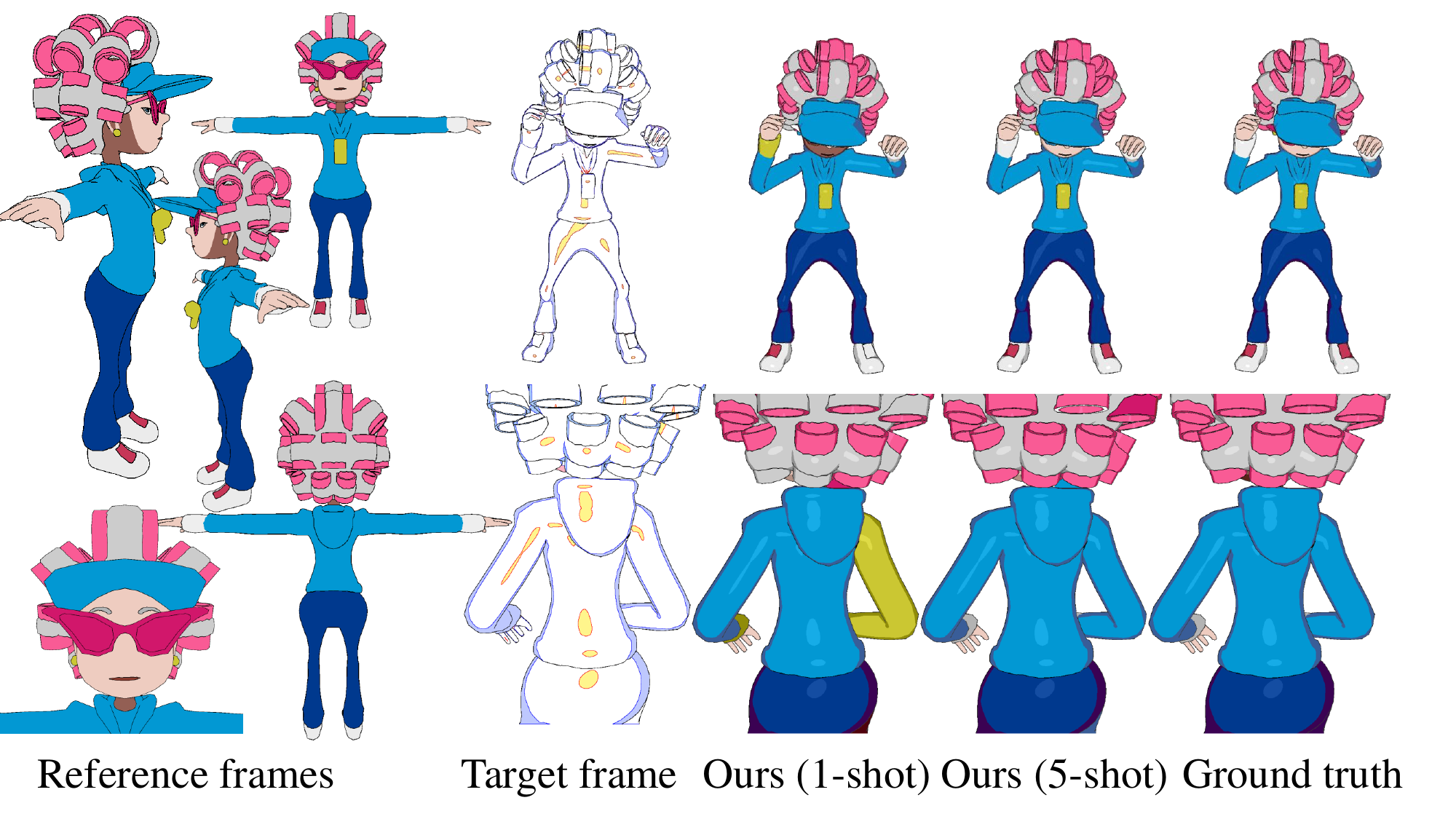}
   \vspace{-5mm}
   \caption{\pami{Visual comparisons of our \textit{keyframe colorization} method using single and multiple references. The front T-pose in the top right serves as the reference for the single-reference estimation. Using multiple references enables the model to achieve more robust results across different perspectives.}}
   \vspace{-4mm}
   \label{fig:few_shot}
\end{figure}

\begin{table*}[]
\vspace{3mm}
\caption{
Quantitative comparison of the impact of different datasets, network structures, and methods for \textit{consecutive frame colorization}. Since the inclusion matching pipeline and coarse warping module with RAFT can also be applied to other networks, we also adopt them to AnT for comparison to illustrate the generalization of these methods. 
} \label{tab:ablation}
\vspace{-3mm}
\resizebox{1.0\textwidth}{!}{
\begin{tabular}{llcccccccccccc}
\hline
                         &                          &  & \multicolumn{5}{c}{3D rendered test set}                                                                                                                  & \multicolumn{1}{l}{} & \multicolumn{5}{c}{Hand-drawn test set}                                                                                                                  \\ \cline{4-8} \cline{10-14} 
\multirow{-2}{*}{Type}   & \multirow{-2}{*}{Method} &  & \multicolumn{1}{c}{Acc}       & \multicolumn{1}{l}{Acc-Thres} & \multicolumn{1}{l}{Pix-Acc}   & \multicolumn{1}{l}{Pix-F-Acc} & \multicolumn{1}{l}{Pix-B-MIoU} & \multicolumn{1}{l}{} & \multicolumn{1}{c}{Acc}       & \multicolumn{1}{l}{Acc-Thres} & \multicolumn{1}{l}{Pix-Acc}   & \multicolumn{1}{l}{Pix-F-Acc} & \multicolumn{1}{l}{Pix-B-MIoU} \\ \cline{1-8} \cline{10-14} 
Baseline                 & Ours                     &  & \textbf{0.8266}               & \textbf{0.8726}               & \textbf{0.9905}               & \textbf{0.9724}               & \textbf{0.9948}                & \textbf{}            & \textbf{0.8593}               & \textbf{0.8929}               & \textbf{0.9900}               & \textbf{0.9638}               & \textbf{0.9984}                \\ \cline{1-2} \cline{4-8} \cline{10-14} 
Dataset                  & AnimeRun                 &  & 0.7359                        & 0.7958                        & 0.9693                        & 0.9201                        & 0.9765                         &                      & 0.8278                        & 0.8634                        & 0.9818                        & 0.9399                        & 0.9906                         \\ \cline{1-2} \cline{4-8} \cline{10-14} 
                         & w/o CLIP                  &  & 0.8180                        & 0.8653                        & 0.9873                        & 0.9685                        & 0.9910                         &                      & 0.8489                        & 0.8843                        & 0.9878                        & 0.9570                        & 0.9925                         \\
                         & w/o RAFT                  &  & 0.8001                        & 0.8447                        & 0.9824                        & 0.9481                        & 0.9878      && 0.8446                        & 0.8750                        & 0.9804                        & 0.9404                        & 0.9851                                                                      \\
\multirow{-3}{*}{Model}  & w/o DConv                 &  &  0.8247 &  0.8685 &  0.9894 &  0.9709 & 0.9928                         &                      &  0.8525 &  0.8833 &  0.9875 &  0.9549 & 0.9937                         \\ \cline{1-2} \cline{4-8} \cline{10-14} 
                         & Ours w/o IM                    &  & 0.7637                        & 0.8352                        & 0.9721                        & 0.9573                        & 0.9688                         &                      & 0.8348                        & 0.8763                        & 0.9777                        & 0.9495                        & 0.9822                         \\
                         & AnT                      &  & 0.7450                        & 0.7930                        & 0.9708                        & 0.9361                        & 0.9758                         &                      & 0.8162                        & 0.8497                        & 0.9817                        & 0.9374                        & 0.9905                         \\
                         & AnT with IM                 &  & 0.8063                        & 0.8447                        & 0.9837                        & 0.9551                        & 0.9887                         &                      & 0.8388                        & 0.8698                        & 0.9839                        & 0.9388                        & 0.9943                         \\

\multirow{-4}{*}{Network} & AnT with IM/RAFT             &  & 0.8177    & 0.8587    & 0.9879    & 0.9681    & 0.9932     &  & 0.8483   & 0.8828   & 0.9886    & 0.9598    & 0.9946     \\ \hline
\end{tabular}
}
\vspace{-3mm}
\end{table*}

\begin{figure*}[t]
  \centering
   \includegraphics[width=0.9\linewidth]{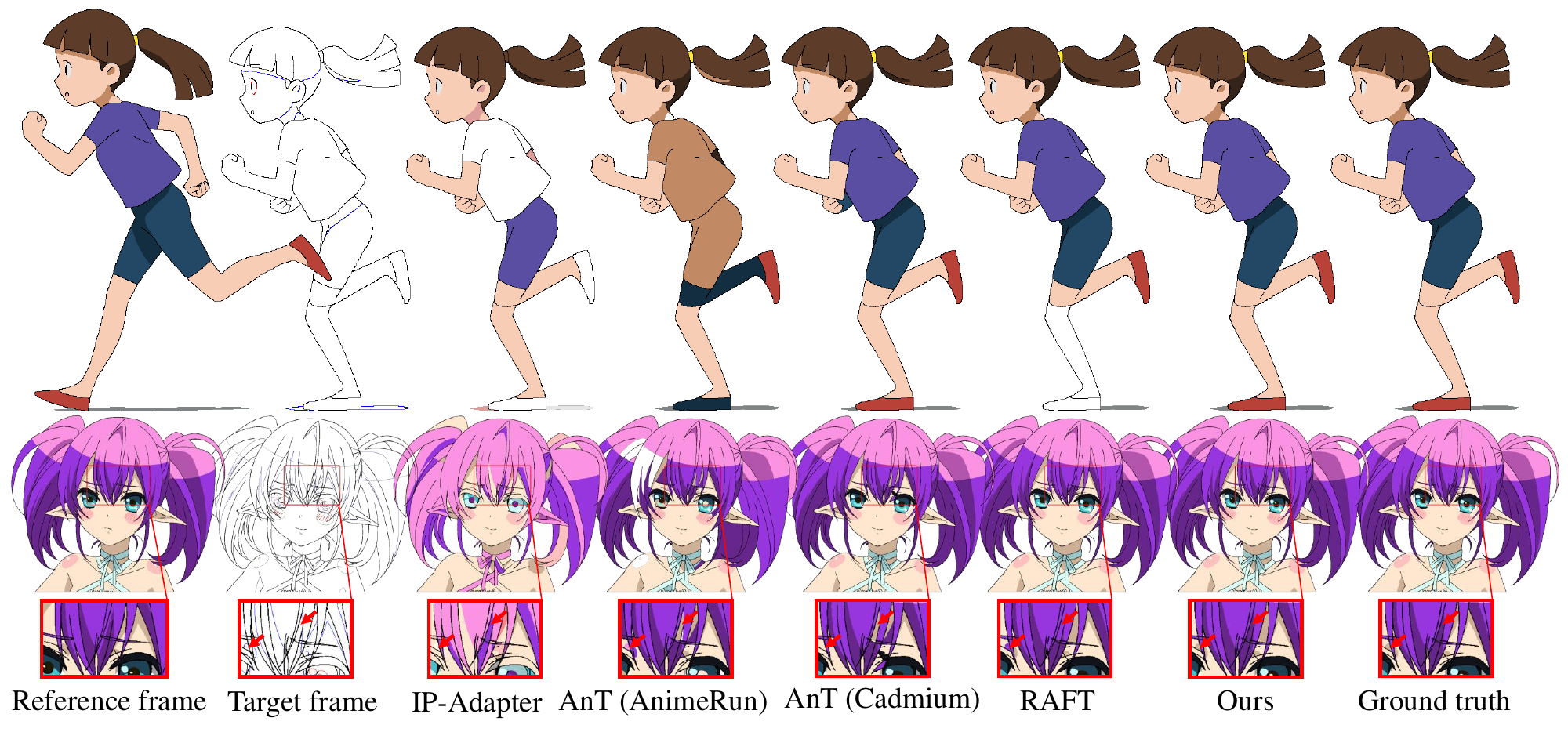}
   \vspace{-4mm}
   \caption{The visualization results of our proposed \textit{consecutive frame colorization} methods and other approaches on real hand-drawn animation. Compared with previous methods, our proposed approach demonstrates enhanced robustness in handling occlusion situations (\eg the hand of the running girl and shadow near the eyebrows in the second row. )}
   \vspace{-3mm}
   \label{fig:real_anime}
\end{figure*}

\begin{figure}[t]
  \centering
   \includegraphics[width=1.0\linewidth]{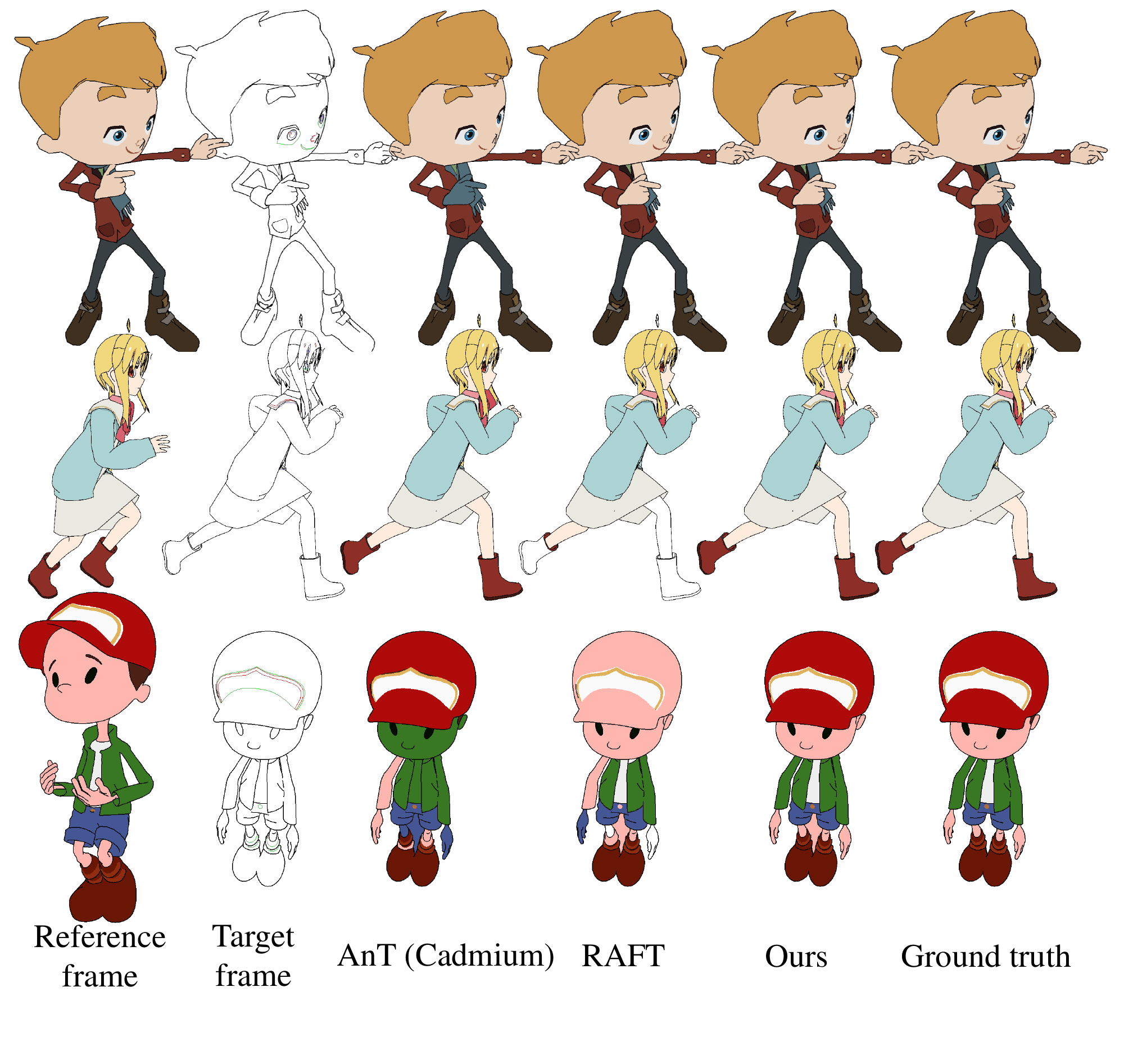}
   \vspace{-10mm}
   \caption{Visual comparison of different \textit{consecutive frame colorization} methods on our 3D rendered test set. Our proposed method produces satisfactory results in the presence of occlusion, significant motion, and substantial changes in viewing angles.}
   \vspace{-4mm}
   \label{fig:synthetic_anime}
\end{figure}

\begin{figure*}[t]
  \centering
   \includegraphics[width=1.0\linewidth]{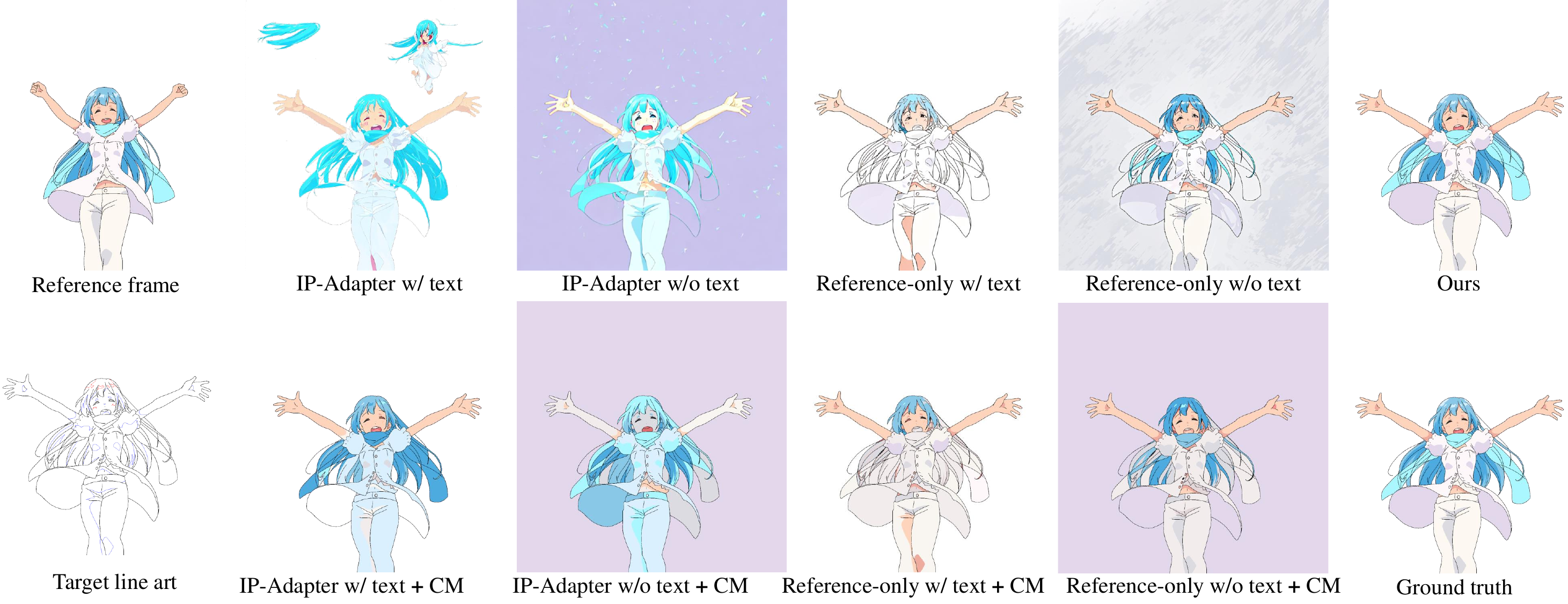}
   \vspace{-6mm}
   \caption{Visual comparisons of our \textit{consecutive frame colorization} method and different reference-based line art colorization methods. `w/ \& w/o text' means using or not using the text prompt `A character in the white background'. In the second row, we employ `CM' (color mapping), a technique that maps the average color in each segment to the closest color in the reference frame. Compared with these reference-based colorization methods, our method can yield more reliable results by avoiding filling similar colors in segments.}
   \vspace{-2mm}
   \label{fig:controlnet}
\end{figure*}

%% file: sec/5_experiments.tex
\begin{figure*}[t]
  \centering
   \includegraphics[width=1.0\linewidth]{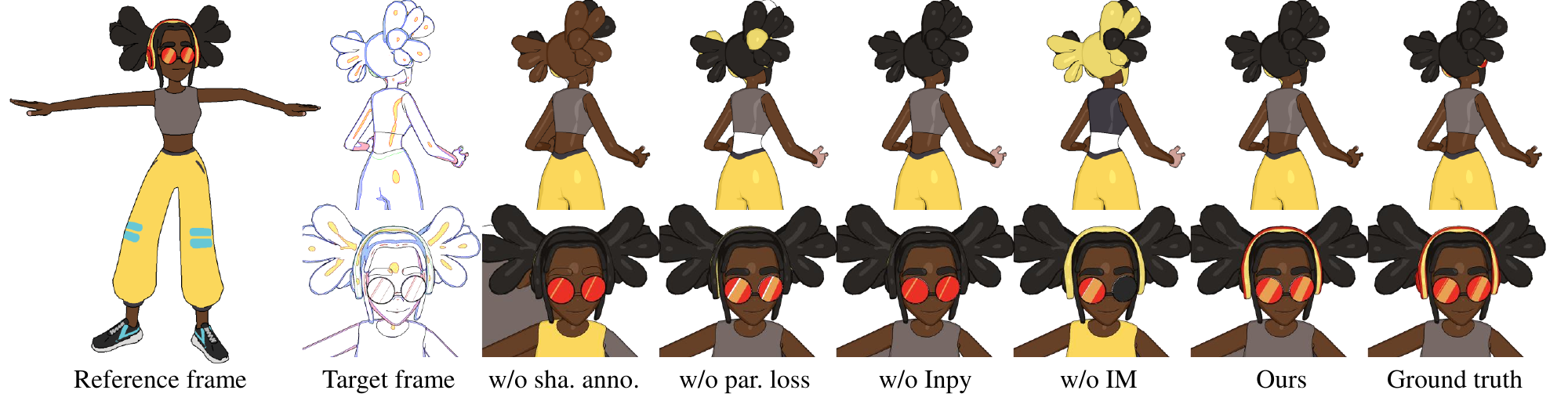}
   \vspace{-8mm}
   \caption{\pami{The Ablation study of different components in our proposed \textit{keyframe colorization} methods.}}
   \vspace{-5mm}
   \label{fig:key_ablation}
\end{figure*}

\begin{figure*}[t]
  \centering
   \includegraphics[width=1.0\linewidth]{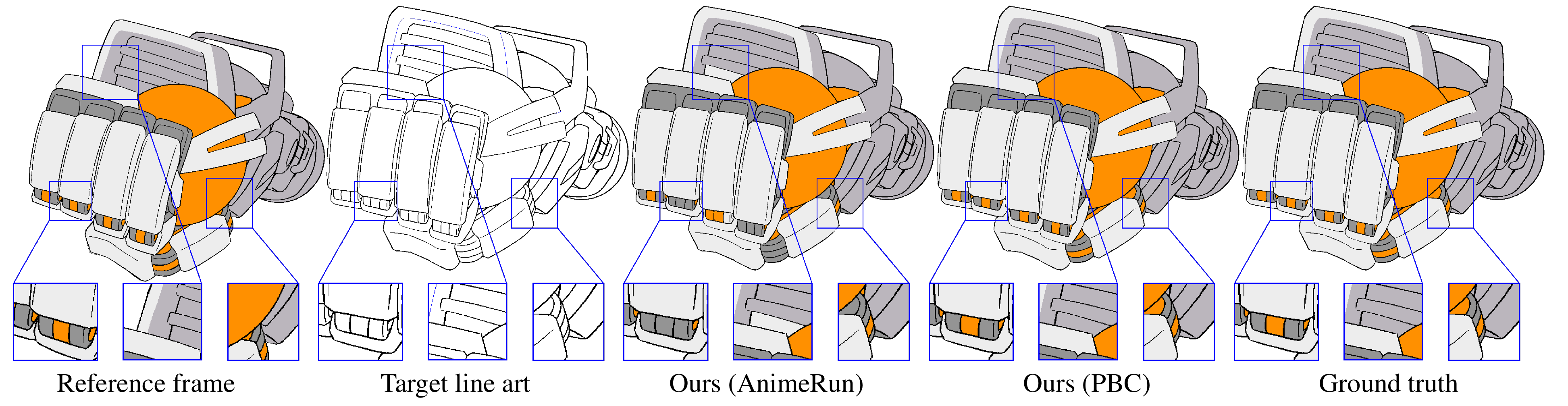}
   \vspace{-6mm}
   \caption{\pami{Visual comparison on training \textit{consecutive frame colorization} network with different datasets. `PBC' represents our proposed dataset \pbc. }}
   \vspace{-3mm}
   \label{fig:ablation_dataset}
\end{figure*}

\begin{figure*}[t]
  \centering
   \includegraphics[width=1.0\linewidth]{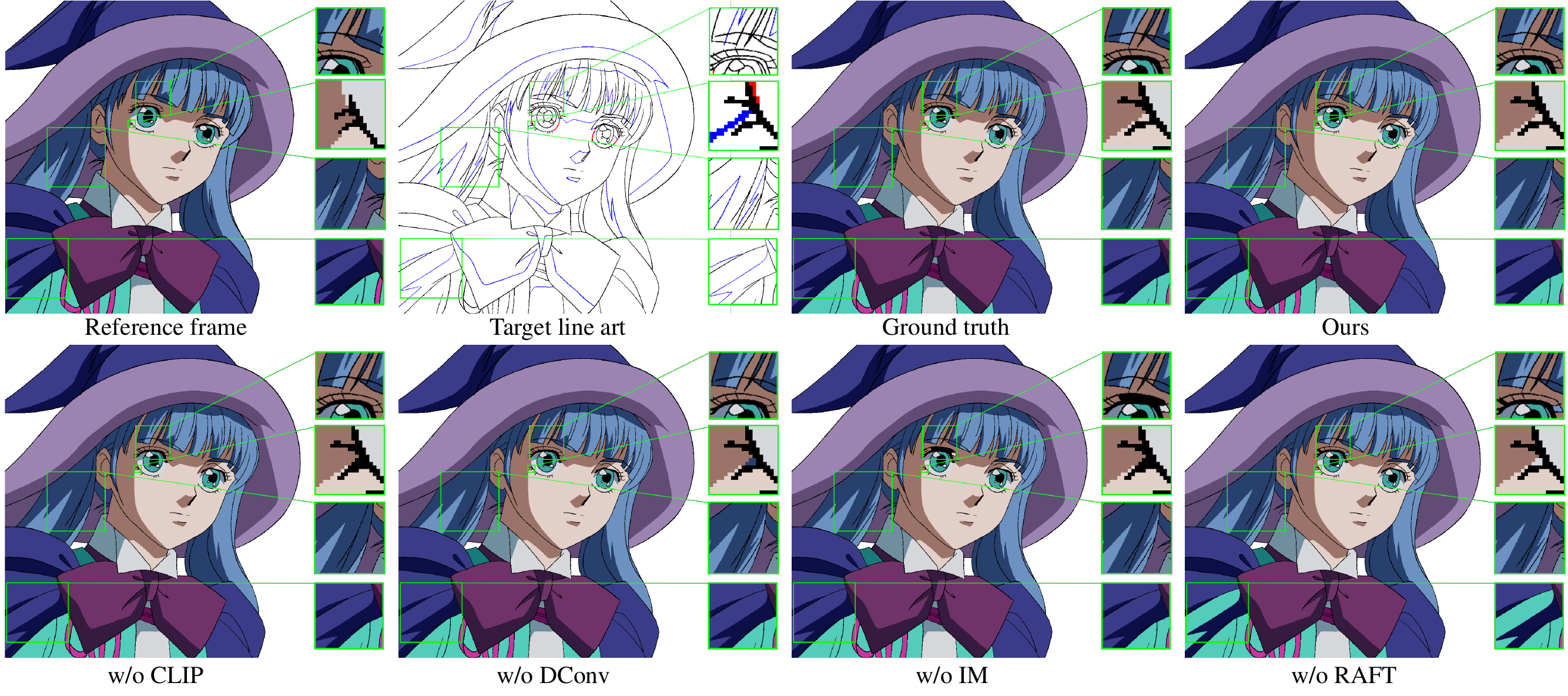}
   \vspace{-7mm}
   \caption{\pami{Ablation study for different modules of our \textit{consecutive frame colorization} method. Zoom in for better visualization.}}
   \vspace{-5mm}
   \label{fig:ablation_study}
\end{figure*}

\begin{figure*}[h]
  \centering
   \includegraphics[width=0.9\linewidth]{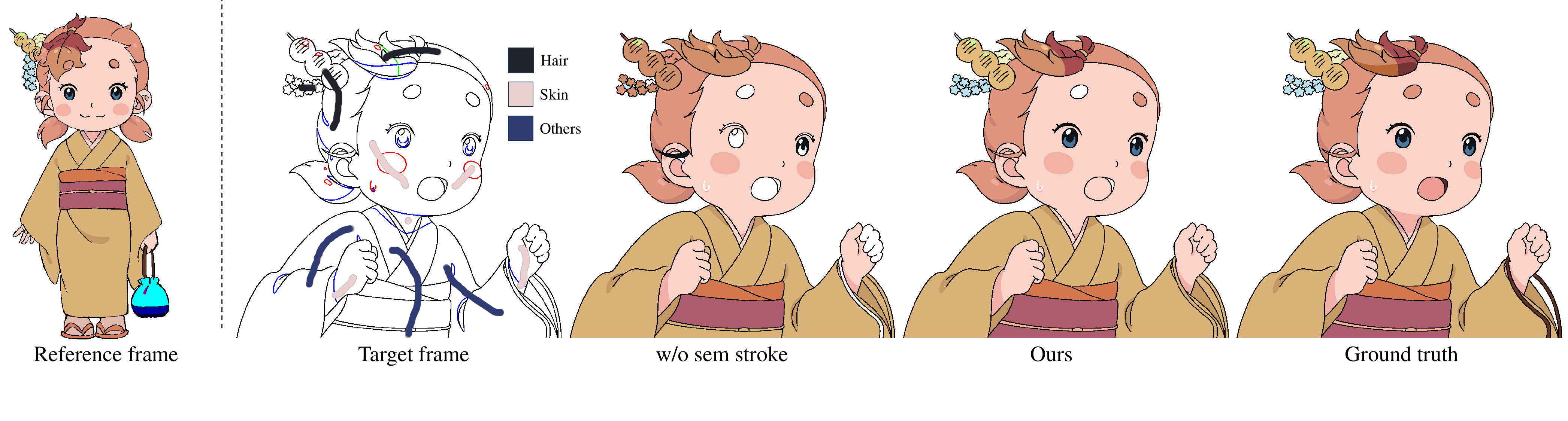}
   \vspace{-9mm}
   \caption{\pami{Illustration of semantic strokes for \textit{keyframe colorization}. Animators can label hair, skin, and other regions with different colors using strokes, even in the absence of shading annotations. These semantic strokes offer a straightforward and efficient interaction method, balancing the workload for animators while enhancing the method's performance. }}
   \vspace{-4mm}
   \label{fig:semantic_stroke}
\end{figure*}

\begin{figure*}[t]
  \centering
   \includegraphics[width=0.9\linewidth]{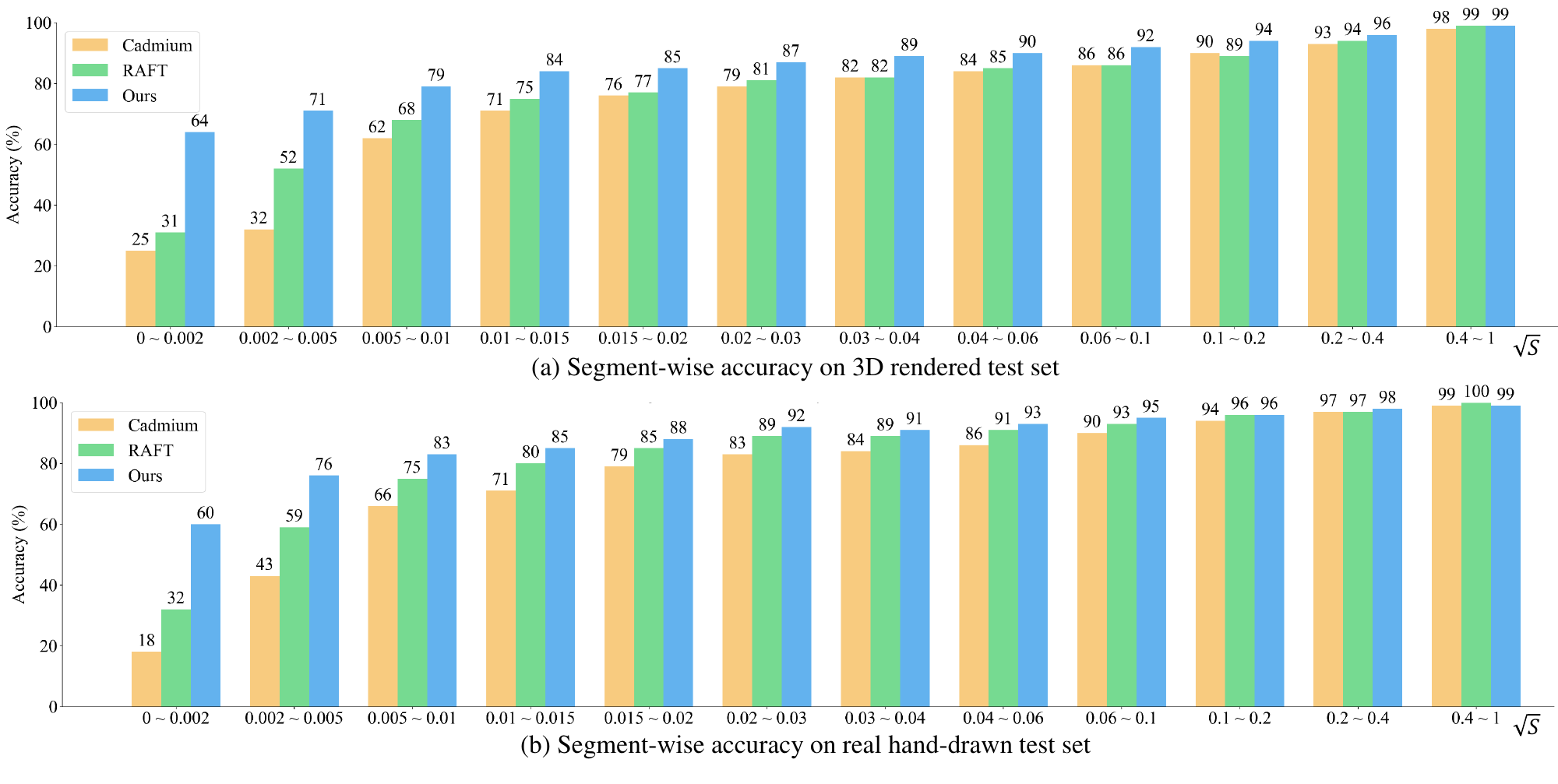}
   \vspace{-5mm}
   \caption{\pami{\textit{Consecutive frame colorization} accuracy comparisons for segments of different sizes. We calculate the segment-wise accuracy in different bins, each partitioned according to the proportional area of the respective segment, denoted as $S$. Given that most segments occupy small areas, we partition the intervals using $\sqrt{S}$ and ensure a roughly equal number of segments in each bin. The RAFT~\cite{teed2020raft} model is finetuned on the AnimeRun~\cite{siyao2022animerun} dataset, and the official implementation of AnT~\cite{AnT} is realized through the Cadmium application~\cite{cadmium}.}}
   \label{fig:seg_accu_label}
\end{figure*}

\begin{figure*}[t]
  \centering
  \vspace{-3mm}
   \includegraphics[width=1.0\linewidth]{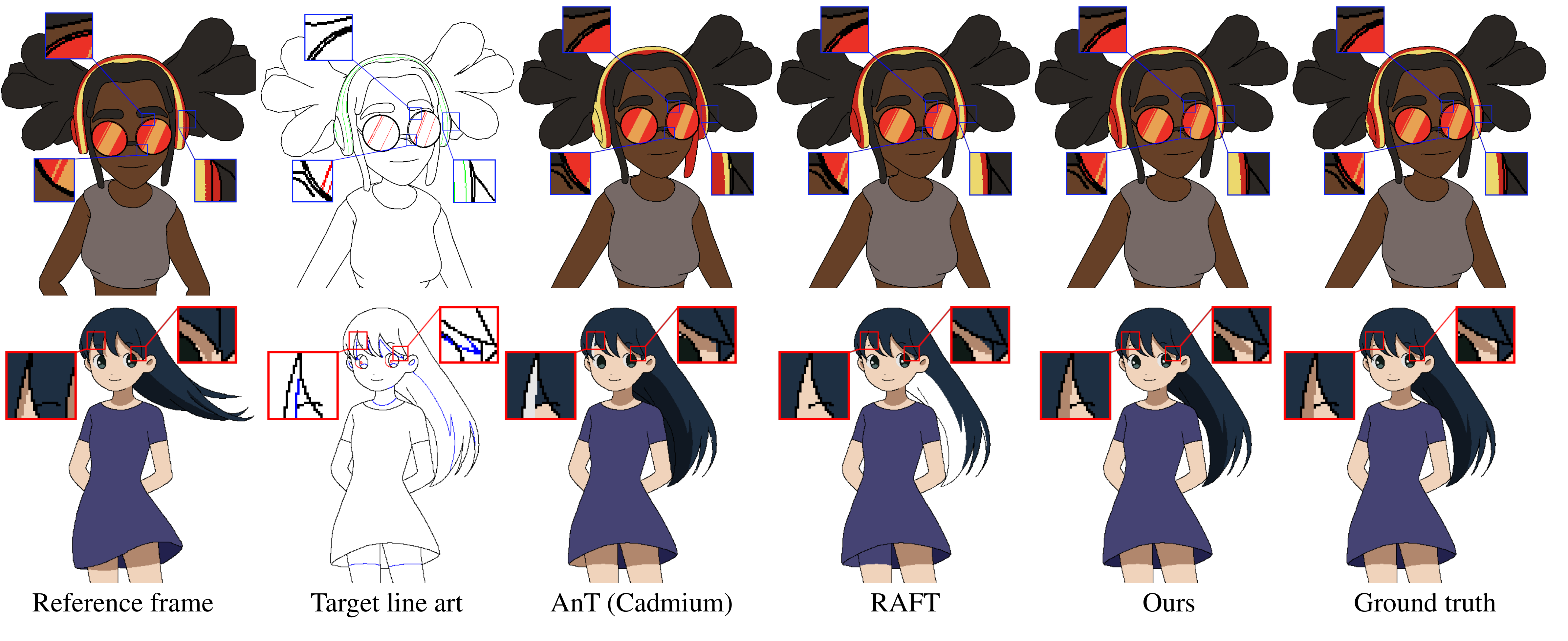}
   \vspace{-8mm}
   \caption{\pami{Visual comparisons of tiny segment colorization across different \textit{consecutive frame colorization} methods. Even in scenarios with no corresponding segment in the reference frame, our approach still excels in colorizing tiny segments accurately.}}
   \vspace{-3mm}
   \label{fig:tiny_seg}
\end{figure*}

\section{Experiments}

\pami{

\subsection{Data Augmentation}
To train the network, we need to sample paired images as reference and target frames from our dataset.
For the keyframe colorization training, we randomly select two frames of each character's whole sequence as reference and target frames.
During the consecutive frame colorization training process, we set the inter-frame intervals of the reference and target frames to 0, 1, and 2.
Notably, even when the reference frame and target frame are the same, the inclusion relationship still may not be apparent to the network.
Additionally, we employ separate random translation in $U(-400,400)$, random rotation in $U(-\pi/{9},\pi/9)$, and random resize in $U(1/2,1/3)$ for paired frames to emulate typical movements in animation. 
In the resizing augmentation, a min pooling strategy is implemented to prevent the removal of black pixels through nearest or bilinear interpolation.
After these augmentation steps, all training data undergo cropping to achieve a consistent size of $640 \times 640$. 
}

\pami{
\subsection{Comparisons with Previous Methods}
Since there is no established method specially designed for keyframe colorization, we primarily compare our \textit{keyframe colorization} method with reference-based line art colorization~\cite{controlnet} and segment matching methods~\cite{AnT,siyao2022animerun}.
We also compare our \textit{consecutive frame colorization} method against the aforementioned methods, as well as optical flow estimation methods~\cite{teed2020raft}.
Detailed descriptions of these approaches are provided below:
}

\begin{enumerate}[itemsep=0pt, topsep=0pt,label=\Alph*)]
\item
\pami{
\noindent{\bf Reference-based Line Art Colorization.}
The saturated luminance of blank regions in line art undermines the effectiveness of gray-scale reference-based colorization methods in line art colorization.
Besides, most reference-based line art colorization methods\cite{chen2020active, lee2020reference,wu2023self,zhang2021line,akita2023hand,xing2024tooncrafter} are not open-source.
Recently, ControlNet~\cite{controlnet} introduces a novel approach to encode the line art of the target frame for reference-based line art colorization.
Finetuning-based methods such as DreamBooth~\cite{dreambooth} and pretrained adapters such as IP-Adapter~\cite{ye2023ip-adapter} serve potential solutions to encode colorized reference frames naturally.
We use Stable Diffusion v1.5~\cite{stable-diffusion} as our base model, employing the DDIM scheduler with 50 steps.
For the finetuning-based method, we finetune the additional LoRA~\cite{lora} for each character based on 10 frames of each character's T-pose in the color design sheet following DreamBooth~\cite{dreambooth}. 
Then, we use ControlNet's line art model and finetuned LoRA to colorize the target line with the text prompt `A [V] character in the white background'.
For the pretrained adapter, we adopt the IP-Adapter~\cite{ye2023ip-adapter} to encode the reference image while using the text prompt 'A character in the white background'.
During the evaluation, we use a color mapping (CM) strategy that assigns the average color in each segment to the closest color in either the reference frame or the text-color list, depending on whether it is consecutive frame colorization or keyframe colorization.
}

\item
\noindent{\bf Segment Matching.}
The official implementation of AnT~\cite{AnT} is available through the Cadmium application~\cite{cadmium}.
Therefore, we primarily compare our method against AnimenRun's reimplementation of AnT~\cite{siyao2022animerun} and the Cadmium application, version 0.3.1.

\item
\noindent{\bf Optical Flow Estimation.}
We choose RAFT~\cite{teed2020raft} as the baseline of our optical flow estimation and compare our consecutive frame colorization method with RAFT finetuned on the AnimeRun dataset.
Since optical flow methods do not inherently provide segment-wise estimations without post-processing, we warp the colors and, during evaluation, assign each segment the most frequently occurring color within it.

\end{enumerate}

\pami{
\noindent{\bf Keyframe Colorization.}
The visual comparisons of different keyframe colorization methods are shown in Fig.~\ref{fig:key_comparison}.
We compare our proposed method with the reference-based methods including the finetuning-based method (LoRA), and pretrained-adapter-based method (IP-Adapter).
Since the Cadmium application does not support the keyframe colorization, we only compare our method with matching-based methods, including AnimeRun's reimplemented AnT (AnimeRun) and our inclusion matching method (BasicPBC), which we introduced in the previous conference version for consecutive frame colorization.
As shown in the results of reference-based methods in Fig.~\ref{fig:key_comparison}, the limited resolution during the denoising process of latent diffusion poses challenges in preserving the fine color details of characters, such as the eyes. Additionally, these reference-based methods may mistakenly apply incorrect but similar-looking colors, as seen with skin tones in IP-Adapter.
Moreover, all these methods struggle to colorize uncommon perspectives accurately. 
Compared with these methods, our method can achieve more accurate results and can even function well on pixel-level details, as shown in the eyes of Fig.~\ref{fig:key_comparison}.
The quantitative comparisons for keyframe colorization, shown in Table~\ref{tab:comparison_ref}, also demonstrate our method's superior accuracy and higher pixel-wise foreground accuracy compared to these approaches.
For multiple reference colorization, we use five images, including a four-view character sheet and face shot as each character's reference frames, as shown in Fig.~\ref{fig:few_shot}.
Compared to using a single reference frame, inference with multiple references improves accuracy across different perspectives.

\noindent{\bf Consecutive Frame Colorization.}
We show the quantitative comparisons for consecutive frame colorization in Table~\ref{tab:baseline}.
We compare our method with various reference-based, segment-matching-based, and optical-flow-based approaches, demonstrating the superior performance of our proposed method.
Since finetuning RAFT can promote the optical flow's performance effectively, we use the finetuned RAFT trained on AnimeRun's results in the following visualizations.
The results of different consecutive frame colorization methods on real hand-drawn animation and rendered 3D test sets are displayed in Fig.~\ref{fig:real_anime} and Fig.~\ref{fig:synthetic_anime}, respectively.
The visual comparisons in Fig.~\ref{fig:real_anime} demonstrate that our method can generalize well on real hand-drawn animation, particularly in scenarios where strict segment correspondence is absent.
Figure~\ref{fig:synthetic_anime} offers a clear visual comparison, demonstrating how our method outperforms recent state-of-the-art segment-matching and optical flow techniques, particularly in scenarios involving occlusions, significant motion, and large perspective changes.
To illustrate the defects of current reference-based methods, more ControlNet-based methods' visualization results are shown in Fig.~\ref{fig:controlnet}.
In the comparisons, we adopt the IP-Adapter~\cite{ye2023ip-adapter} and ControlNet's reference-only preprocessor~\cite{controlnet} to encode the reference image, and test using or not using the text prompt and color mapping strategy.
These comparative results demonstrate that current reference-based methods still cannot restore accurate colors even with the color mapping strategy.
}

\pami{
\subsection{Ablation Study for Keyframe Colorization}

\noindent{\bf Model.} To examine whether each architecture design enhances colorization accuracy, we conduct experiments by removing each key component: the input pyramid of the feature extraction network (w/o Inpy), the parsing loss in the segment parsing module (w/o par. loss), the shading annotations used during training (w/o sha. anno.), and inclusion matching training pipeline (w/o IM). 
In the experiment of w/o IM, we follow the matching strategy in Fig.~\ref{fig:inclusion_matching}(a) and match the index label with the index label instead of doing inclusion matching.
As shown in Table~\ref{tab:ablation_ref} and Fig.~\ref{fig:key_ablation}, both shading annotations and parsing loss effectively guide the network in correctly colorizing characters from various perspectives. The face shot in the second row of Fig.~\ref{fig:key_ablation} demonstrates how the input pyramid enables our method to colorize details in the line art at different scales. Furthermore, the inclusion matching pipeline significantly enhances the visual quality of the results.

\noindent{\bf Network.} To demonstrate the superiority of our keyframe colorization network, we also train AnT~\cite{AnT} and BasicPBC~\cite{InclusionMatching2024} using our dataset and inclusion matching pipeline.
The quantitative comparison in Table~\ref{tab:ablation_ref} highlights the effectiveness of our proposed method.
Since all the methods are trained with randomly selected 2 frames from each sequence, previous matching-based approaches struggle to match segments from different perspectives, leading to lower accuracy compared to pretrained official models.
}

\subsection{Ablation Study for Consecutive Frame Colorization}
\noindent{\bf Dataset.} To demonstrate the effectiveness of our proposed dataset, we also train our proposed method on the AnimeRun. In the training process, we adopt the inclusion matching strategy by merging the segments in the reference frames. The quantitative results reported in Table~\ref{tab:ablation} and visual comparison in Fig.~\ref{fig:ablation_dataset} suggest the positive influence of our proposed dataset.
\pami{Although our dataset only consists of anime characters, Fig.~\ref{fig:ablation_dataset} also demonstrates that our consecutive frame colorization methods can generalize well to different scenes.}

\noindent{\bf Model.} 
To examine the necessity of each structure, we explore alternative configurations by replacing the deformable convolution and offset estimation module with the original convolution (w/o DConv), eliminating the color warping module (w/o RAFT), applying segment matching instead of inclusion matching (w/o IM), and omitting the CLIP image encoder for ablation. 
The results presented in Table~\ref{tab:ablation} underscore the significance of each design. 
The visual comparisons in Fig.~\ref{fig:ablation_study} particularly highlight the crucial role of the color warping module and inclusion matching. 
The warped color feature introduces a leakage of grouping information from the reference frame, facilitating the feature extraction network in learning inclusion relationships more effectively compared to relying solely on urging the multiplex transformer to acquire this representation.
Thus, lacking coarse warping or inclusion matching will all lead to mismatching while there is no one-to-one correspondence, as shown in the eyelash part of Fig.~\ref{fig:ablation_study}.

\noindent{\bf Network.} The results in Table~\ref{tab:ablation} further highlight the efficacy of our inclusion matching pipeline. Its application to AnT~\cite{AnT} yields a substantial improvement in network performance, suggesting the broad applicability and generalization of our inclusion matching pipeline.

\pami{
\subsection{Semantic Stroke for Keyframe Colorization.}
In our keyframe colorization pipeline, we assume that all keyframe animations adhere to professional animation production standards, which include shading annotations.
For animations that lack these annotations, animators can still use strokes with different colors to label regions of hair, skin, and others, as illustrated in Fig.~\ref{fig:semantic_stroke}. We refer to these as 'semantic strokes.'
Highlight and shadow regions are identified by locating the smallest areas adjacent to the color lines.
As shown in Fig.~\ref{fig:semantic_stroke}, semantic strokes offer a simple and efficient way to balance the animator's workload with the colorization result.

\subsection{Consecutive Frame Colorization for Tiny Segments}
Colorizing tiny segments poses a significant challenge in paint bucket colorization, consuming a considerable amount of time for digital painters. Despite the existence of methods based on segment matching~\cite{AnT} and optical flow, accurate colorization of tiny segments remains challenging.
In Fig.~\ref{fig:seg_accu_label}, we evaluate the accuracy of segment colorization across various sizes. Our proposed method outperforms current optical-flow-based approaches~\cite{teed2020raft,siyao2022animerun} and AnT~\cite{AnT}, particularly excelling in tiny segments.
Even when dealing with segments as small as a few pixels, the proposed inclusion matching demonstrates its effectiveness in guiding the network to produce accurate and reliable results. 

}

%% file: sec/6_limitation.tex
\pami{
\section{Limitation}
Although our dataset and method significantly improve the performance of paint bucket colorization, several limitations remain.
For keyframe colorization, the performance is hindered on smaller segments due to imperfections in the segment parsing module, often requiring digital painters to fix erroneous colors manually.
Additionally, the accuracy of our multiple reference colorization method relies heavily on selecting appropriate reference frames for matching. Simply increasing the number of reference frames does not consistently improve accuracy, and in some cases, choosing unsuitable frames may result in worse outcomes.
For consecutive frame colorization, our method still requires that all segments in the target frame also appear in the reference frame, making it difficult to accurately colorize new segments in dynamic scenes, such as when characters open their eyes or turn around.
Besides, while our inclusion matching pipeline effectively handles one-to-many and many-to-many matching scenarios, it may struggle with complex occlusions, exaggerated motion, or distortions, potentially resulting in incorrect matches. 
Despite these challenges, our method demonstrates superior performance compared to existing approaches, even in these demanding scenarios.
}

%% file: sec/7_conclusion.tex
\pami{
\section{Conclusion}
Line art colorization is an essential part of animation production. Digital painters colorize these line arts based on the character color design sheet. To achieve automatic colorization, we developed a systematic solution comprising a keyframe colorization method based on the character color design sheet and a consecutive frame colorization method for color propagation.
Departing from segment matching methods that struggle with one-to-one correspondence, we propose a novel inclusion matching pipeline that estimates each segment's inclusion relationship rather than relying on direct correspondence. To facilitate the learning of this inclusion relationship, we introduce a dedicated dataset, \pbc, and implement two distinct colorization pipelines tailored to different tasks.
Our keyframe colorization pipeline includes a segment parsing component, while the consecutive colorization pipeline integrates a coarse color warping module. Both pipelines share a common inclusion matching module, ensuring consistent and accurate colorization across frames with occlusion and perspective changes.
Experimental results demonstrate the superior performance of our method in handling occlusion and accommodating large movements within challenging scenarios.
}

%% file: main.bbl
\begin{thebibliography}{10}
\providecommand{\url}[1]{#1}
\csname url@samestyle\endcsname
\providecommand{\newblock}{\relax}
\providecommand{\bibinfo}[2]{#2}
\providecommand{\BIBentrySTDinterwordspacing}{\spaceskip=0pt\relax}
\providecommand{\BIBentryALTinterwordstretchfactor}{4}
\providecommand{\BIBentryALTinterwordspacing}{\spaceskip=\fontdimen2\font plus
\BIBentryALTinterwordstretchfactor\fontdimen3\font minus \fontdimen4\font\relax}
\providecommand{\BIBforeignlanguage}[2]{{%
\expandafter\ifx\csname l@#1\endcsname\relax
\typeout{** WARNING: IEEEtran.bst: No hyphenation pattern has been}%
\typeout{** loaded for the language `#1'. Using the pattern for}%
\typeout{** the default language instead.}%
\else
\language=\csname l@#1\endcsname
\fi
#2}}
\providecommand{\BIBdecl}{\relax}
\BIBdecl

\bibitem{InclusionMatching2024}
Y.~Dai, S.~Zhou, Q.~Li, C.~Li, and C.~C. Loy, ``Learning inclusion matching for animation paint bucket colorization,'' \emph{CVPR}, 2024.

\bibitem{sykora2004unsupervised}
D.~S{\`y}kora, J.~Buri{\'a}nek, and J.~{\v{Z}}{\'a}ra, ``Unsupervised colorization of black-and-white cartoons,'' in \emph{International Symposium on Non-photorealistic Animation and Rendering}, 2004.

\bibitem{sykora2009rigid}
D.~S{\`y}kora, J.~Dingliana, and S.~Collins, ``As-rigid-as-possible image registration for hand-drawn cartoon animations,'' in \emph{International Symposium on Non-photorealistic Animation and Rendering}, 2009.

\bibitem{chen2020active}
S.-Y. Chen, J.-Q. Zhang, L.~Gao, Y.~He, S.~Xia, M.~Shi, and F.-L. Zhang, ``Active colorization for cartoon line drawings,'' \emph{IEEE TVCG}, vol.~28, no.~2, 2020.

\bibitem{lee2020reference}
J.~Lee, E.~Kim, Y.~Lee, D.~Kim, J.~Chang, and J.~Choo, ``Reference-based sketch image colorization using augmented-self reference and dense semantic correspondence,'' in \emph{CVPR}, 2020.

\bibitem{wu2023self}
S.~Wu, X.~Yan, W.~Liu, S.~Xu, and S.~Zhang, ``Self-driven dual-path learning for reference-based line art colorization under limited data,'' \emph{IEEE TCSVT}, vol.~34, no.~3, 2023.

\bibitem{zhang2021line}
Q.~Zhang, B.~Wang, W.~Wen, H.~Li, and J.~Liu, ``Line art correlation matching feature transfer network for automatic animation colorization,'' in \emph{WACV}, 2021.

\bibitem{controlnet}
L.~Zhang, A.~Rao, and M.~Agrawala, ``Adding conditional control to text-to-image diffusion models,'' in \emph{ICCV}, 2023.

\bibitem{akita2023hand}
K.~Akita, Y.~Morimoto, and R.~Tsuruno, ``Hand-drawn anime line drawing colorization of faces with texture details,'' \emph{Computer Animation and Virtual Worlds}, 2023.

\bibitem{cao2023animediffusion}
Y.~Cao, X.~Meng, P.~Mok, X.~Liu, T.-Y. Lee, and P.~Li, ``Animediffusion: Anime face line drawing colorization via diffusion models,'' \emph{IEEE TVCG}, vol.~30, no.~10, 2024.

\bibitem{xing2024tooncrafter}
J.~Xing, H.~Liu, M.~Xia, Y.~Zhang, X.~Wang, Y.~Shan, and T.-T. Wong, ``Tooncrafter: Generative cartoon interpolation,'' \emph{arXiv preprint arxiv:2405.17933}, 2024.

\bibitem{AnT}
E.~Casey, V.~P{\'e}rez, and Z.~Li, ``The animation transformer: Visual correspondence via segment matching,'' in \emph{ICCV}, 2021.

\bibitem{siyao2022animerun}
L.~Siyao, Y.~Li, B.~Li, C.~Dong, Z.~Liu, and C.~C. Loy, ``Animerun: 2d animation visual correspondence from open source 3d movies,'' \emph{NeurIPS}, 2022.

\bibitem{maejima2019graph}
A.~Maejima, H.~Kubo, T.~Funatomi, T.~Yotsukura, S.~Nakamura, and Y.~Mukaigawa, ``Graph matching based anime colorization with multiple references,'' in \emph{SIGGRAPH}, 2019.

\bibitem{maejima2024continual}
A.~Maejima, S.~Shinagawa, H.~Kubo, T.~Funatomi, T.~Yotsukura, S.~Nakamura, and Y.~Mukaigawa, ``Continual few-shot patch-based learning for anime-style colorization,'' \emph{CVM}, vol.~10, no.~4, 2024.

\bibitem{kyoto_draw}
K.~A. publishing department, \emph{The Kyoto Animation Guide to Drawing}.\hskip 1em plus 0.5em minus 0.4em\relax Kyoto Animation Co., Ltd, 2010.

\bibitem{mixamo}
``Mixamo,'' \url{https://www.mixamo.com/}.

\bibitem{aplaybox}
``Aplaybox,'' \url{https://www.aplaybox.com/}.

\bibitem{kim_tag2pix_2019}
H.~Kim, H.~Y. Jhoo, E.~Park, and S.~Yoo, ``{Tag2Pix}: Line art colorization using text tag with {SECat} and changing loss,'' in \emph{ICCV}, 2019.

\bibitem{zouSA2019sketchcolorization}
C.~Zou, H.~Mo, C.~Gao, R.~Du, and H.~Fu, ``Language-based colorization of scene sketches,'' \emph{ACM TOG}, vol.~38, no.~6, 2019.

\bibitem{sykora2009lazybrush}
D.~S{\`y}kora, J.~Dingliana, and S.~Collins, ``Lazybrush: Flexible painting tool for hand-drawn cartoons,'' in \emph{CGF}, vol.~28, no.~2, 2009.

\bibitem{xia-2018-invertible}
L.~Zhang, C.~Li, T.-T. Wong, Y.~Ji, and C.~Liu, ``Two-stage sketch colorization,'' \emph{ACM TOG}, vol.~37, no.~6, 2018.

\bibitem{Filling2021zhang}
L.~Zhang, C.~Li, E.~Simo-Serra, Y.~Ji, T.-T. Wong, and C.~Liu, ``User-guided line art flat filling with split filling mechanism,'' in \emph{CVPR}, 2021.

\bibitem{cao2021line}
R.~Cao, H.~Mo, and C.~Gao, ``Line art colorization based on explicit region segmentation,'' in \emph{CGF}, vol.~40, 2021.

\bibitem{huang2024lvcd}
Z.~Huang, M.~Zhang, and J.~Liao, ``Lvcd: reference-based lineart video colorization with diffusion models,'' in \emph{SIGGRAPH}, 2024.

\bibitem{siyao_deep_2021}
L.~Siyao, S.~Zhao, W.~Yu, W.~Sun, D.~Metaxas, C.~C. Loy, and Z.~Liu, ``Deep animation video interpolation in the wild,'' in \emph{CVPR}, 2021.

\bibitem{sintel}
D.~J. Butler, J.~Wulff, G.~B. Stanley, and M.~J. Black, ``A naturalistic open source movie for optical flow evaluation,'' in \emph{ECCV}, 2012.

\bibitem{shugrina2019creative}
M.~Shugrina, Z.~Liang, A.~Kar, J.~Li, A.~Singh, K.~Singh, and S.~Fidler, ``Creative flow+ dataset,'' in \emph{CVPR}, 2019.

\bibitem{liu_shape_2023}
S.~Liu, X.~Wang, X.~Liu, Z.~Wu, and H.~S. Seah, ``Shape correspondence for cel animation based on a shape association graph and spectral matching,'' \emph{CVM}, 2023.

\bibitem{liu2020shape}
S.~Liu, X.~Wang, Z.~Wu, and H.~S. Seah, ``Shape correspondence based on kendall shape space and rag for 2d animation,'' \emph{The Visual Computer}, vol.~36, 2020.

\bibitem{zhu-2016-toontrack}
H.~Zhu, X.~Liu, T.-T. Wong, and P.-A. Heng, ``Globally optimal toon tracking,'' \emph{ACM TOG}, vol.~35, no.~4, 2016.

\bibitem{zhang_excol_2012}
L.~Zhang, H.~Huang, and H.~Fu, ``{EXCOL}: An {EXtract}-and-{COmplete} layering approach to cartoon animation reusing,'' \emph{IEEE TVCG}, vol.~18, no.~7, 2012.

\bibitem{dang_correspondence_2020}
T.~D.~Q. Dang, T.~Do, A.~Nguyen, V.~Pham, Q.~Nguyen, B.~Hoang, and G.~Nguyen, ``Correspondence neural network for line art colorization,'' in \emph{{SIGGRAPH}}, 2020.

\bibitem{hu1962visual}
M.-K. Hu, ``Visual pattern recognition by moment invariants,'' \emph{IRE transactions on information theory}, vol.~8, no.~2, 1962.

\bibitem{unet}
O.~Ronneberger, P.~Fischer, and T.~Brox, ``U-net: Convolutional networks for biomedical image segmentation,'' in \emph{MICCAI}, 2015.

\bibitem{vaswani2017attention}
A.~Vaswani, N.~Shazeer, N.~Parmar, J.~Uszkoreit, L.~Jones, A.~N. Gomez, {\L}.~Kaiser, and I.~Polosukhin, ``Attention is all you need,'' \emph{NeurIPS}, 2017.

\bibitem{sarlin20superglue}
P.-E. Sarlin, D.~DeTone, T.~Malisiewicz, and A.~Rabinovich, ``{SuperGlue}: Learning feature matching with graph neural networks,'' in \emph{CVPR}, 2020.

\bibitem{chen2020simclr}
T.~Chen, S.~Kornblith, M.~Norouzi, and G.~Hinton, ``A simple framework for contrastive learning of visual representations,'' in \emph{ICML}, 2020.

\bibitem{he2020moco}
K.~He, H.~Fan, Y.~Wu, S.~Xie, and R.~Girshick, ``Momentum contrast for unsupervised visual representation learning,'' in \emph{CVPR}, 2020.

\bibitem{clip}
A.~Radford, J.~W. Kim, C.~Hallacy, A.~Ramesh, G.~Goh, S.~Agarwal, G.~Sastry, A.~Askell, P.~Mishkin, J.~Clark \emph{et~al.}, ``Learning transferable visual models from natural language supervision,'' in \emph{ICML}, 2021.

\bibitem{jia2021scaling}
C.~Jia, Y.~Yang, Y.~Xia, Y.-T. Chen, Z.~Parekh, H.~Pham, Q.~Le, Y.-H. Sung, Z.~Li, and T.~Duerig, ``Scaling up visual and vision-language representation learning with noisy text supervision,'' in \emph{ICML}, 2021.

\bibitem{reda2022film}
F.~Reda, J.~Kontkanen, E.~Tabellion, D.~Sun, C.~Pantofaru, and B.~Curless, ``Film: Frame interpolation for large motion,'' in \emph{ECCV}, 2022.

\bibitem{zhao2017pyramid}
H.~Zhao, J.~Shi, X.~Qi, X.~Wang, and J.~Jia, ``Pyramid scene parsing network,'' in \emph{CVPR}, 2017.

\bibitem{zhou2022lednet}
S.~Zhou, C.~Li, and C.~Change~Loy, ``{LEDNet}: Joint low-light enhancement and deblurring in the dark,'' in \emph{ECCV}, 2022.

\bibitem{teed2020raft}
Z.~Teed and J.~Deng, ``Raft: Recurrent all-pairs field transforms for optical flow,'' in \emph{ECCV}, 2020.

\bibitem{dai2017deformable}
J.~Dai, H.~Qi, Y.~Xiong, Y.~Li, G.~Zhang, H.~Hu, and Y.~Wei, ``Deformable convolutional networks,'' in \emph{ICCV}, 2017.

\bibitem{cherti2023reproducible}
M.~Cherti, R.~Beaumont, R.~Wightman, M.~Wortsman, G.~Ilharco, C.~Gordon, C.~Schuhmann, L.~Schmidt, and J.~Jitsev, ``Reproducible scaling laws for contrastive language-image learning,'' in \emph{CVPR}, 2023.

\bibitem{cadmium}
``Cadmium,'' \url{https://cadmium.app/}.

\bibitem{dreambooth}
N.~Ruiz, Y.~Li, V.~Jampani, Y.~Pritch, M.~Rubinstein, and K.~Aberman, ``Dreambooth: Fine tuning text-to-image diffusion models for subject-driven generation,'' in \emph{CVPR}, 2023.

\bibitem{ye2023ip-adapter}
H.~Ye, J.~Zhang, S.~Liu, X.~Han, and W.~Yang, ``{IP}-adapter: Text compatible image prompt adapter for text-to-image diffusion models,'' \emph{arXiv preprint arxiv:2308.06721}, 2023.

\bibitem{stable-diffusion}
R.~Rombach, A.~Blattmann, D.~Lorenz, P.~Esser, and B.~Ommer, ``High-resolution image synthesis with latent diffusion models,'' in \emph{CVPR}, 2022.

\bibitem{lora}
E.~J. Hu, Y.~Shen, P.~Wallis, Z.~Allen-Zhu, Y.~Li, S.~Wang, L.~Wang, and W.~Chen, ``Lo{RA}: Low-rank adaptation of large language models,'' in \emph{ICLR}, 2022.

\end{thebibliography}
